\documentclass[11pt]{article}

\usepackage[final]{acl}

\usepackage{times}
\usepackage{latexsym}

\usepackage[T1]{fontenc}
\usepackage[utf8]{inputenc}

\usepackage{microtype}

\usepackage{inconsolata}

\usepackage{graphicx}
\usepackage{subcaption}
\usepackage{multirow}
\usepackage{booktabs} % for professional tables
\usepackage[table,dvipsnames]{xcolor}
\usepackage{tabularx}
\usepackage{array}

\usepackage{amsmath}
\usepackage{amssymb}
\usepackage{mathtools}
\usepackage{amsthm}
\usepackage{threeparttable}
\usepackage{adjustbox}
\usepackage{algorithm}
\usepackage{algpseudocode}

\algrenewcommand{\algorithmicrequire}{\textbf{Input:}}
\algrenewcommand{\algorithmicensure}{\textbf{Output:}}
\algrenewcommand{\algorithmiccomment}[1]{%
    \hfill$\triangleright$ \textcolor{gray}{#1}}

\title{S$^4$R: Selective Sampling, Subspaces, and Sparse Reconstruction for Compressed Long-Context KV Caching}

\author{
 \textbf{Jialong Han},
 \textbf{You Wu},
 \textbf{Kewei Tu}\thanks{Corresponding author.}
 \\
School of Information Science and Technology, ShanghaiTech University\\
Shanghai Engineering Research Center of Intelligent Vision and Imaging
\\
 \texttt{\{hanjl2026, wuyou2024, tukw\}@shanghaitech.edu.cn}
}

\begin{document}
\maketitle
\begin{abstract}
  The growth of context window lengths in Large Language Models (LLMs) significantly enhances their long-context capabilities but incurs prohibitive memory costs due to the Key-Value (KV) cache.
  Although low-rank compression of KV cache is a promising remedy, existing methods face a dilemma: offline approaches depend on external calibration data, whereas online approaches incur substantial compute for full-prompt decomposition and reconstruction.
  In this paper, we propose \textbf{S$^4$R}, which builds low-rank subspaces from selectively sampled tokens and computes attention over a sparsely reconstructed KV representation.
  S$^4$R uses prompt-aware initialization to build initial key/value bases from a representative prompt subset, trading off calibration-data dependence against prefilling cost.
  Because fully reconstructing the cache at every decoding step is prohibitively expensive and hurts throughput, we further adopt sparse reconstruction to retain only informative positions during decoding.
  Extensive experiments on LongBench and RULER with Llama and Qwen model families show that S$^4$R achieves up to \textbf{5$\times$} KV compression with near full-cache accuracy, combining the efficiency of fixed compression with the adaptability of prompt-dependent methods.
  Our code is publicly available at \url{https://github.com/Draconis98/S4R}
\end{abstract}

\section{Introduction}
The remarkable expansion of context window sizes in Large Language Models (LLMs) --- now spanning from 128k up to millions of tokens \cite{gpt42024, gemini2024,qwen251m2025} --- has ushered in a new era of capabilities, powering breakthroughs in long-document summarization \cite{bai-etal-2025-longbench}, retrieval-augmented generation \cite{retrievalaugmented2020}, and autonomous agents \cite{generative2023,survey2024a}. 
The scaling of context window brings an acute bottleneck: memory consumption. As context length increases, the Key-Value (KV) cache --- used for storing intermediate computations --- can rapidly surpass the footprint of the model parameters themselves \cite{efficient2023,MLSYS2023_c4be71ab, ainslie-etal-2023-gqa}, making memory efficiency the central challenge in LLMs inference.

To address this challenge, numerous KV cache compression techniques have been explored, 
among which low-rank methods --- particularly Singular Value Decomposition (SVD) --- stand out for their effectiveness \cite{survey2025b}. 
These approaches are motivated by the empirical observation that the KV cache often exhibits strong low-rank structures \cite{effectively2025}, which can be exploited to reduce memory usage. However, existing low-rank methods face a trade-off between efficiency and adaptivity. 
Fixed post-training methods either derive projection bases from calibration activations or factorize KV projection weights before inference, and then reuse the resulting low-dimensional representations during serving \cite{palu2025,saxena-etal-2024-eigen,lorc2024,recalkv2025}. 
They avoid per-prompt decomposition, but their accuracy can depend on calibration-data match and rank allocation, particularly under aggressive compression \cite{shadowkv2025,commonkv2025,zhu-etal-2026-ojakv}. 
% Cross-layer schemes exploit an orthogonal source of redundancy by sharing KV structure across layers \cite{commonkv2025}. 
Prompt-dependent methods estimate or adapt low-rank representations during inference, which improves robustness to the current context but introduces additional computation, especially during prefilling \cite{shadowkv2025,xkv2025,zhu-etal-2026-ojakv}.

Our key insight is that the dominant KV subspace of a long prompt need not be estimated from either an external calibration set or the entire prompt. 
Instead, it can be approximated from a carefully selected subset of prompt tokens. 
At the same time, KV cache entries should not be compressed or reconstructed uniformly. 
Sink tokens serve as stable global pivots in attention and should be preserved in full precision \cite{efficient2024}, while most non-sink tokens can be stored in a low-rank coefficient space and reconstructed only when they are likely to matter for the current decoding step. 

Based on this insight, we propose \textbf{S$^4$R}, a prompt-aware low-rank KV cache compression method with selective sampling and sparse reconstruction. 
During prefilling, S$^4$R keeps sink tokens in full precision, samples representative non-sink tokens from the prompt, normalizes their KV states, and constructs key/value subspaces with truncated SVD. 
It then stores non-sink keys and values as low-rank coefficients rather than full-dimensional KV states. During decoding, S$^4$R estimates token relevance directly in the latent coefficient space, combines a local window with globally important positions, and reconstructs only the selected KV entries for final attention computation. 
This design combines the low prefilling overhead of fixed compression with the context adaptivity of prompt-dependent methods.

We evaluate S$^4$R on LongBench \cite{bai-etal-2024-longbench} and RULER \cite{ruler2024} with Llama \cite{llama2024} and Qwen \cite{qwen251m2025,qwen252025,yang2025qwen3technicalreport} model families.
On LongBench, S$^4$R achieves up to \textbf{5$\times$} KV cache compression and remains close to the full-cache baseline across model scales. 
Across accuracy and efficiency evaluations, S$^4$R consistently matches or outperforms prior compressed-cache baselines while substantially reducing long-context inference latency.

\begin{itemize}
  \item We propose S$^4$R, a prompt-aware low-rank KV cache compression method that builds compact key/value subspaces from selectively sampled prompt tokens and combines sink-token preservation, latent-space relevance estimation, and sparse reconstruction to reduce persistent KV memory while preserving context adaptivity.
  \item We evaluate S$^4$R on LongBench and RULER, showing near full-cache accuracy under up to \textbf{5$\times$} KV compression and a favorable accuracy--efficiency trade-off against strong compressed-cache baselines.
\end{itemize}

\section{Related Work}
Low-rank KV cache compression reduces redundancy along the hidden or channel dimension, and is complementary to methods that shorten the cached sequence, quantize cache entries, merge tokens, or move them across the memory hierarchy \cite{survey2025b}. The most relevant low-rank methods differ along three axes: where the compression subspace is obtained, how rank is allocated across keys, values, heads, or layers, and whether decoding requires reconstructing the full cache.

\paragraph{Pre-computed subspaces and adaptive rank allocation.}
Several methods determine most of the low-rank representation before serving the current prompt. Palu and LoRC factorize key/value projection weights and cache compact intermediate states, with LoRC using progressive layer-wise compression to control accumulated error \cite{palu2025,lorc2024}. Eigen Attention derives bases from calibration activations and performs attention in the induced low-dimensional space, while ReCalKV treats keys and values separately through head reordering, grouped SVD, offline calibration, and matrix fusion \cite{saxena-etal-2024-eigen,recalkv2025}. CSKV adds lightweight reconstruction training and a bi-branch cache that preserves a recent full-precision window \cite{cskv2024}. KQ-SVD instead chooses the projection by directly minimizing query--key inner-product distortion and gives a closed-form low-rank solution for preserving pre-softmax attention scores \cite{kqsvd2026}. Other work adapts the compression granularity: TALE assigns token-dependent ranks and avoids eager reconstruction \cite{lee-etal-2025-tale}, STAR-KV learns fine-grained ranks through differentiable singular-value thresholding \cite{starkv2026}, and CommonKV uses SVD-based parameter sharing with adaptive cross-layer budget allocation \cite{commonkv2025}. These approaches reduce or remove prompt-time decomposition, but their bases, rank policies, or calibration statistics are not constructed solely from the current prompt.

\paragraph{Prompt-adaptive subspaces.}
Prompt-dependent methods align the compressed representation with the current context, at the cost of online decomposition or updates. ShadowKV applies SVD to current-prompt pre-RoPE keys, stores a low-rank key cache on GPU, offloads values, and reconstructs selected sparse KV pairs during decoding \cite{shadowkv2025}. xKV performs prompt-time cross-layer SVD and shares aligned low-rank structure across grouped layers \cite{xkv2025}. OjaKV continuously adapts its basis with online PCA, using a comprehensive prefilling update, lightweight periodic decoding updates, and full-rank anchors \cite{zhu-etal-2026-ojakv}. S$^4$R follows this context-adaptive direction, but constructs both key and value bases from a sampled subset of non-sink prompt tokens rather than decomposing or updating over the full prompt.

\paragraph{Latent-space selection and sparse reconstruction.}
Low-rank representations can also serve as inexpensive proxies for deciding which tokens merit full attention. Loki uses compressed keys to approximate attention scores for sparse token selection, but does not itself reduce the persistent value cache \cite{loki2024}. SALS stores a compressed latent cache, scores tokens through RoPE-free query--key interactions in that space, and reconstructs only a selected subset \cite{sals}. S$^4$R shares this reconstruction-aware design, while coupling it with prompt-sampled key/value bases, full-precision sink tokens, a local query window, and a persistent coefficient cache for both keys and values. It therefore occupies the intersection of prompt-adaptive low-rank compression and latent-space sparse reconstruction. Appendix~\ref{sec:appendix_related_work} covers complementary eviction, merging, quantization, and system-oriented methods.
% and provides finer-grained comparisons with representative baselines.

\section{Method}

S$^4$R consists of two main stages: the \textit{prefilling} and the 
\textit{decoding} stages.
We defer background on standard KV cache computation and SVD to Appendix~\ref{sec:appendix_preliminaries}.

\begin{algorithm}[t]
\caption{S$^4$R Prefilling Stage}
\label{alg:prefilling}
\begin{algorithmic}[1]

\Require Non-sink pre-RoPE states
         $\mathbf{S}^K,\mathbf{S}^V$;
         recent query window $\widetilde{\mathbf{Q}}$;
         sampling budget $T^\prime$, and rank $r$;
         $H\in\{K,V\}$

\Ensure Low-rank bases $\mathbf{B}^K,\mathbf{B}^V$,
        coefficient caches $\mathbf{C}^K,\mathbf{C}^V$,
        and query coefficient window $\mathbf{C}^Q$

\State $\mathcal{I}
       \gets \Call{HybridSample}{T,T^\prime}$

\State $\mathbf{S}_{[\mathcal{I}]}^H
       \gets
       \Call{Normalize}{
           \mathbf{S}_{[\mathcal{I}]}^H,
           \mathrm{dim}=-1
       }$

\State $(\mathbf{U}^H,\mathbf{\Sigma}^H,\mathbf{B}^H)
       \gets
       \operatorname{SVD}_{r}
       \left(\mathbf{S}_{[\mathcal{I}]}^H\right)$

\State $\mathbf{C}^H
       \gets
       \mathbf{S}^H\mathbf{B}^H$
       \Comment{Project all non-sink states}

\State $\mathbf{C}^Q
       \gets
       \widetilde{\mathbf{Q}}\mathbf{B}^K$
       \Comment{Initialize the latent query window}

\State \Return
       $\mathbf{B}^K,\mathbf{B}^V,
        \mathbf{C}^K,\mathbf{C}^V,\mathbf{C}^Q$

\end{algorithmic}
\end{algorithm}

\subsection{Prefilling Stage}\label{sec:method_prefill}
The main motivation of the prefilling stage is to efficiently construct suitable low-rank subspaces for keys and values. 
As directly applying SVD to the full long-context sequence (e.g., 128k tokens) is computationally prohibitive, this stage starts by selecting a representative subset of tokens from the input prompt to approximate the dominant subspace structure of the full KV states.
Together with the full-precision preservation of sink tokens,
Algorithm~\ref{alg:prefilling} summarizes the prompt-aware
subspace construction and coefficient-cache initialization
performed during prefilling.

\subsubsection{Sink Token Handling}
Due to their role as global pivots in attention, sink tokens typically exhibit consistently high attention weights across queries and layers \cite{efficient2024}. 
As a result, their representations are critical for maintaining stable long-range dependencies and cannot be reliably approximated by a low-rank subspace without incurring noticeable information loss.

To preserve their fidelity, we exclude the first $s$ sink tokens from the compression pipeline and store their pre-RoPE key and value representations in full precision.
% These tokens are always included during attention computation without reconstruction. 
The remaining tokens are treated as non-sink tokens and are subject to low-rank approximation.

\subsubsection{Subspace Construction}\label{sec:token_selection}

Let the non-sink tokens in the input prompt be denoted as $\mathbf{X}\in\mathbb{R}^{\mathcal{B}\times \mathcal{T}\times \mathcal{D}}$. 
We construct a subset of tokens $\mathbf{X}_{[\mathcal{I}]}\in\mathbb{R}^{\mathcal{B}\times \mathcal{T}'\times \mathcal{D}}$ by selecting $\mathcal{T}'\ll\mathcal{T}$ using a hybrid sampling strategy that combines (i) uniform sampling over early positions and (ii) a contiguous block of most recent (right) tokens. 
Here, $\mathcal{I}\in\mathbb{N}^{\mathcal{B}\times \mathcal{T}'}$ denotes the indices of the selected non-sink tokens. 
Correspondingly, the non-sink tokens are partitioned into two disjoint subsets: the selected tokens ($\mathbf{X}_{[\mathcal{I}]}$) and the remaining tokens ($\mathbf{X}_{[\widetilde{\mathcal{I}}]}$), where $\widetilde{\mathcal{I}}$ denotes the complement of $\mathcal{I}$ along the sequence dimension.

Before applying SVD to the selected tokens, we perform token-wise normalization to mitigate the influence of varying vector magnitudes on the learned subspace. 
In practice, tokens with larger norms may dominate the decomposition and bias the resulting singular vectors toward high-energy directions. 
To alleviate this effect, we normalize each sampled pre-RoPE key and value representation along the feature dimension:
\begin{equation}
  \mathbf{S}^{K/V}_{[\mathcal{I}]}\leftarrow\Call{normalize}{\mathbf{S}^{K/V}_{[\mathcal{I}]},\text{dim=}-1}.
\end{equation}

This normalization step ensures that the subspace primarily captures directional information rather than being skewed by magnitude differences, leading to a more stable and representative low-rank approximation.

Then, we perform truncated SVD for each layer:
\begin{equation}
  \mathbf{U}^{K/V}, \mathbf{\Sigma}^{K/V}, \mathbf{B}^{K/V} \leftarrow \mathrm{SVD}_r\left(\mathbf{S}^{K/V}_{[\mathcal{I}]}\right),
  \label{eq:svd}
\end{equation}
where $\mathbf{U}^{K/V} \in \mathbb{R}^{\mathcal{B} \times \mathcal{T}' \times r}$ are the left singular vectors, $\mathbf{\Sigma}^{K/V} \in \mathbb{R}^{\mathcal{B} \times r \times r}$ are the singular values, and $\mathbf{B}^{K/V} \in \mathbb{R}^{\mathcal{B} \times \mathcal{D}' \times r}$ are the top-$r$ right singular vectors, which serve as the low-rank bases.

We then project the non-sink key and value representations onto the learned subspaces to obtain their coefficient representations:
\begin{equation}
  \mathbf{C}^{K/V} = \mathbf{S}^{K/V}\mathbf{B}^{K/V},
  \label{eq:coefficients}
\end{equation}
where $\mathbf{C}^{K/V}\in\mathbb{R}^{\mathcal{B}\times \mathcal{T}\times r}$ are the projection coefficients.

\subsubsection{Initial Local Query Window}
To enable efficient relevance estimation without reconstructing the full keys and values during decoding, we also project the local queries into the same low-rank subspace.

Let the most recent queries be denoted as $\widetilde{\mathbf{Q}}\in\mathbb{R}^{\mathcal{B}\times\mathcal{G}\times\mathcal{W}\times\mathcal{D}'}$.
The projected query coefficients are then computed as
\begin{equation}
  \mathbf{C}^Q=\widetilde{\mathbf{Q}}\mathbf{B}^{K}
\end{equation}
where $\mathbf{C}^Q\in\mathbb{R}^{\mathcal{B}\times\mathcal{G}\times\mathcal{W}\times r}$.

\begin{algorithm}[t]
\caption{S$^4$R Decoding at Step $t$}
\label{alg:decoding}
\begin{algorithmic}[1]

\Require New pre-RoPE states
         $\mathbf{s}_t^Q,\mathbf{s}_t^K,\mathbf{s}_t^V$;
         bases $\mathbf{B}^K,\mathbf{B}^V$;
         coefficient caches
         $\mathbf{C}_{t-1}^K,\mathbf{C}_{t-1}^V$;
         query window $\mathbf{C}^Q$;
         sink states $\mathbf{K}_{\mathrm{sink}},
         \mathbf{V}_{\mathrm{sink}}$;
         $W$, $r$, $\kappa$, and $\rho$; $H\in\{K,V\}$

\Ensure Attention output $\mathbf{O}_t$ and updated
        $\mathbf{C}_t^K,\mathbf{C}_t^V,\mathbf{C}^Q$

\State $\mathbf{c}_t^{Q/K}
       \gets \mathbf{s}_t^{Q/K}\mathbf{B}^K$
\State $\mathbf{c}_t^{V}
        \gets \mathbf{s}_t^{V}\mathbf{B}^V$
       \Comment{Project the new states}

\State $\mathbf{C}_t^H
       \gets
       \mathbf{C}_{t-1}^H \oplus \mathbf{c}_t^H$
       
       \Comment{Append to the coefficient cache}

\State $\mathbf{C}^Q
       \gets
       \Call{concat}{
           \mathbf{C}^Q[:, :, 1:, :],
           \mathbf{c}_t^Q
       }$
       
       \Comment{Update the FIFO query window}

\State $L_t \gets T+t$

\State $\mathbf{A}
       \gets
       \sigma_{L_t}
       \left(
           \operatorname{Mask}_{\mathrm{causal}}
           (\dfrac{
           \mathbf{C}^Q
           \left(\mathbf{C}_t^K\right)^\top
       }{\sqrt{r}})
       \right)$

\State $\widetilde{\mathbf{A}}
       \gets
       \mathcal{F}_{P}
       \left(\dfrac{1}{W}
       \displaystyle\sum_{i=1}^{W}
       \mathbf{A}[:, :, i, :];\kappa\right)$
       \Comment{Pooling}

\State $\mathcal{I}_{\mathrm{local}}
       \gets
       \{L_t-W,\ldots,L_t-1\}$

\State $k
       \gets
       \max\left(
           \left\lfloor\rho L_t\right\rfloor-W,
           0
       \right)$

\State $\mathcal{I}_{\mathrm{global}}
       \gets
       \Call{TopK}{
           \widetilde{\mathbf{A}}[:, :L_t-W],
           k
       }$

\State $\widehat{\mathcal{I}}
       \gets
       \mathcal{I}_{\mathrm{local}}
       \cup
       \mathcal{I}_{\mathrm{global}}$

\State $\widehat{\mathbf{H}}
       \gets
       {\mathbf{C}_t^H}_{[\widehat{\mathcal{I}}]}
       \left(\mathbf{B}^{H}\right)^\top$

       \Comment{Reconstruct only selected positions}

\State $\mathbf{H}^{\star}
       \gets
       \Call{concat}{
           \mathbf{H}_{\mathrm{sink}},
           \widehat{\mathbf{H}}
       }$

\State $\mathbf{O}_t
       \gets
       \sigma
       \left(
           \dfrac{
               \operatorname{RoPE}(\mathbf{s}_t^Q)
               \operatorname{RoPE}(\mathbf{K}^{\star})^\top
           }{\sqrt{d_h}}
       \right)
       \mathbf{V}^{\star}$
       
       \Comment{Use the original token positions}

\State \Return
       $\mathbf{O}_t,
        \mathbf{C}_t^K,
        \mathbf{C}_t^V,
        \mathbf{C}^Q$

\end{algorithmic}
\end{algorithm}

\subsection{Decoding Stage}\label{sec:method_decode}
At a high level, decoding in S$^4$R keeps the KV cache compressed while reconstructing only a small set of useful tokens at each step.
Each new token is projected into the prefilling-stage low-rank subspace to update the coefficient cache.
Recent query coefficients are then used to score cached tokens in latent space, which provides an inexpensive estimate of token relevance.
S$^4$R reconstructs only the selected local and globally important non-sink tokens, concatenates them with the full-precision sink tokens, and performs attention over this compact set.
Algorithm~\ref{alg:decoding} summarizes the per-step S$^4$R
decoding procedure, including latent-space relevance estimation,
token selection, and sparse KV reconstruction.

\subsubsection{Compressed Cache Update}
Let $\mathbf{s}_t^{Q} \in \mathbb{R}^{\mathcal{B}\times \mathcal{G}\times 1\times \mathcal{D}'}$ and $\mathbf{s}_t^{K/V} \in \mathbb{R}^{\mathcal{B}\times 1\times \mathcal{D}'}$ denote the query, key, and value representations at step $t$, respectively. The corresponding coefficient representations are computed as:
\begin{equation}
  \begin{aligned}
    \mathbf{c}_t^{Q/K} &= \mathbf{s}_t^{Q/K}\mathbf{B}^{K},\\
    \mathbf{c}_t^{V} &= \mathbf{s}_t^{V}\mathbf{B}^{V},
  \end{aligned}
\end{equation}
where $\mathbf{c}_t^{Q}\in\mathbb{R}^{\mathcal{B}\times \mathcal{G}\times 1\times r}$ and $\mathbf{c}_t^{K/V}\in\mathbb{R}^{\mathcal{B}\times 1\times r}$.

The key and value coefficients are appended to the cached sequences:
\begin{equation}
  \mathbf{C}_t^{K/V} = \mathbf{C}_{t-1}^{K/V} \oplus \mathbf{c}_t^{K/V},
\end{equation}
where $\mathbf{C}_t^{K/V}\in\mathbb{R}^{\mathcal{B}\times (\mathcal{T}+t)\times r}$ stores the accumulated coefficients.

In contrast, query coefficients are maintained as a fixed-size sliding window of length $\mathcal{W}$, forming a first-in-first-out queue $\mathbf{C}^Q \in \mathbb{R}^{\mathcal{B}\times \mathcal{G}\times \mathcal{W}\times r}$.
At each step $t$, the new coefficient $\mathbf{c}_t^Q$ is appended, while the oldest entry is removed:
\begin{equation}
  \mathbf{C}^Q \leftarrow
  \Call{concat}{\mathbf{C}^Q[:,:,1:,:],; \mathbf{c}_t^Q}.
\end{equation}

\subsubsection{Latent-Space Relevance Estimation}
To avoid explicit reconstruction of the full key and value representations, we approximate the attention scores directly in the latent coefficient space.

Let $\mathcal{L}_t=\mathcal{T}+t$ denote the current non-sink cache length. Given the query coefficients $\mathbf{C}^Q\in\mathbb{R}^{\mathcal{B}\times\mathcal{G}\times\mathcal{W}\times r}$ and the cached key coefficients $\mathbf{C}_t^K\in\mathbb{R}^{\mathcal{B}\times\mathcal{L}_t\times r}$, we first compute the latent-space attention logits:
\begin{equation}
  \mathbf{Z} = \frac{\mathbf{C}^Q(\mathbf{C}_t^K)^\top}{\sqrt{r}},
\end{equation}
where the inner product is taken along the latent dimension $r$, yielding $\mathbf{Z}\in\mathbb{R}^{\mathcal{B}\times\mathcal{G}\times\mathcal{W}\times\mathcal{L}_t}$.

We then apply the causal mask before normalization, so each query only attends to valid historical positions:
\begin{equation}
  \mathcal{A} = \sigma_{\mathcal{L}_t}\left(\mathrm{Mask}_{\text{causal}}(\mathbf{Z})\right),
\end{equation}
where the softmax ($\sigma$) is computed over the cache-length dimension.

To obtain a single relevance score for each cached token, we average the attention scores over the query window:
\begin{equation}
  \mathcal{A}^{Q} = \frac{1}{\mathcal{W}} \sum_{i=1}^{\mathcal{W}} \mathcal{A}_{[:, :, i, :]}.
\end{equation}
Here, $\mathcal{A}^{Q}\in\mathbb{R}^{\mathcal{B}\times\mathcal{G}\times\mathcal{L}_t}$.

\subsubsection{Sparse Reconstruction and Attention}
Given the relevance scores $\mathcal{A}^{Q} \in \mathbb{R}^{\mathcal{B}\times \mathcal{G}\times\mathcal{L}_t}$, 
we first apply a pooling operation along the sequence dimension before selecting global tokens. 
This follows SnapKV-style token clustering \cite{snapkv2024}, where neighboring tokens are smoothed or grouped to produce more stable importance estimates:
\begin{equation}
  \widetilde{\mathcal{A}}=\mathcal{F}_{\mathcal{P}}(\mathcal{A}^{Q};\kappa)
\end{equation}
where $\mathcal{F}_{\mathcal{P}}$ denotes either average pooling or max pooling with kernel size $\kappa$.

For models with $\mathcal{G} > 1$, we further aggregate across the group dimension using $\mathcal{F}_\mathcal{G}\in\{\mathrm{mean}, \mathrm{max}\}$:
\begin{equation}
  \widetilde{\mathcal{A}} = \mathcal{F}_\mathcal{G}(\widetilde{\mathcal{A}}),
\end{equation}
where $\widetilde{\mathcal{A}} \in \mathbb{R}^{\mathcal{B}\times \mathcal{L}_t}$ represents the final relevance score for each cached token.

We select a subset of historical tokens for reconstruction by combining a local window with globally important tokens.
We always retain the most recent $\mathcal{W}$ tokens:
\begin{equation}
  \mathcal{I}_{\text{local}} = \{\mathcal{L}_t-\mathcal{W}, \dots, \mathcal{L}_t-1\}.
\end{equation}
From the remaining tokens, we select the top-$k$ positions according to $\widetilde{\mathcal{A}}$:
\begin{equation}
  \mathcal{I}_{\text{global}} = \Call{TopK}{\widetilde{\mathcal{A}}_{[:, :\mathcal{L}_t-\mathcal{W}]},\, k},    
\end{equation}
where $k = \max(\lfloor \rho \mathcal{L}_t\rfloor - \mathcal{W} ,\, 0)$ is determined by the reconstruction ratio $\rho$.

The final index set is given by:
\begin{equation}
  \widehat{\mathcal{I}} = \mathcal{I}_{\text{local}} \cup \mathcal{I}_{\text{global}}.
\end{equation}

Given the selected indices, we reconstruct the corresponding key and value representations from the low-rank coefficients:
\begin{equation}
  \widehat{\mathbf{K}} = \mathbf{C}_t^K{}_{[\widehat{\mathcal{I}}]} (\mathbf{B}^K)^\top,\quad
  \widehat{\mathbf{V}} = \mathbf{C}_t^V{}_{[\widehat{\mathcal{I}}]} (\mathbf{B}^V)^\top,
\end{equation}
where $\widehat{\mathbf{K}}, \widehat{\mathbf{V}} \in \mathbb{R}^{\mathcal{B}\times |\widehat{\mathcal{I}}|\times \mathcal{D}'}$.

These reconstructed tokens are then concatenated with the sink tokens:
\begin{equation}
  \begin{aligned}
    \mathbf{K}^\star &= \Call{concat}{\mathbf{K}_{\text{sink}}, \widehat{\mathbf{K}}}, \\
    \mathbf{V}^\star &= \Call{concat}{\mathbf{V}_{\text{sink}}, \widehat{\mathbf{V}}}.
  \end{aligned}
\end{equation}

We then apply RoPE to $\mathbf{K}^\star$ according to their original positions to ensure positional consistency.

The final attention output is computed as:
\begin{equation}
  \mathbf{O}_t = \sigma\left(
  \frac{\text{RoPE}(\mathbf{s}_t^Q) \text{RoPE}(\mathbf{K}^\star)^\top}{\sqrt{d_h}}
  \right)\mathbf{V}^\star.
\end{equation}

\subsection{Memory Analysis}

We analyze the per-layer memory footprint in number of stored scalar elements, omitting the batch dimension for clarity.
At decoding step $t$, the full KV cache stores both keys and values for the $s$ sink tokens and the $\mathcal{L}_t$ non-sink tokens:
\begin{equation}
  M_{\mathrm{full}}(t) = 2(s+\mathcal{L}_t)\mathcal{D}'.
\end{equation}

In S$^4$R, sink tokens are kept in full precision, while non-sink tokens are stored in the low-rank coefficient cache.
The persistent cache therefore consists of the full sink KV states, the key/value coefficient caches, the query-coefficient window, and the learned key/value bases:
\begin{equation}
  M_{\mathrm{S^4R}}(t)
  = 2s\mathcal{D}' + 2\mathcal{L}_t r + \mathcal{G}\mathcal{W}r + 2\mathcal{D}'r.
\end{equation}
The term $\mathcal{G}\mathcal{W}r$ stores the fixed-size query-coefficient window, while the last term stores $\mathbf{B}^{K}$ and $\mathbf{B}^{V}$.
The resulting persistent memory ratio is
\begin{equation}
  \frac{M_{\mathrm{S^4R}}(t)}{M_{\mathrm{full}}(t)}
  =
  \frac{s\mathcal{D}' + \mathcal{L}_t r + \frac{1}{2}\mathcal{G}\mathcal{W}r + \mathcal{D}'r}
  {(s+\mathcal{L}_t)\mathcal{D}'}.
  \label{eq:memory_ratio}
\end{equation}
When $\mathcal{L}_t \gg s,r$, the ratio approaches $r/\mathcal{D}'$, showing that the long-context memory is dominated by the low-rank coefficient dimension.

During attention computation, S$^4$R reconstructs only the selected non-sink tokens.
Let
\begin{equation}
  \begin{aligned}
    m_t = |\widehat{\mathcal{I}}| &= \mathcal{W} + \max(\lfloor \rho\mathcal{L}_t\rfloor-\mathcal{W},0)\\
    &= \max(\mathcal{W}, \lfloor \rho\mathcal{L}_t\rfloor)
  \end{aligned}
\end{equation}
denote the number of reconstructed non-sink tokens.
The transient working memory is then
\begin{equation}
  M_{\mathrm{work}}(t) = 2m_t\mathcal{D}',
\end{equation}
which corresponds to the reconstructed key/value buffer.
Thus, the peak per-step memory is approximately
\begin{equation}
  M_{\mathrm{peak}}(t)
  \approx M_{\mathrm{S^4R}}(t) + M_{\mathrm{work}}(t).
\end{equation}
Since $m_t\ll \mathcal{L}_t$ and $r\ll \mathcal{D}'$, S$^4$R reduces the persistent KV cache substantially while avoiding full-cache reconstruction during decoding.

\section{Experiments}
In this section, we comprehensively evaluate the effectiveness and memory savings of S$^4$R across multiple benchmarks.
We also conduct speed evaluations to analyze the impact of our method on generation efficiency.
Additionally, we perform ablation studies to analyze the contributions of key components within the S$^4$R method.

Unless otherwise specified, the rank $r$ is obtained by solving Eq.~(\ref{eq:memory_ratio}) for the target retention ratio $\eta$, and the reconstruction ratio is set to $\rho=\eta$.
We use $s=4$ sink tokens following the default setting of \citet{efficient2024}, a query-window size of $\mathcal{W}=32$, and a pooling kernel size of $\kappa=7$ following the default setting of \citet{snapkv2024}.
For token relevance aggregation, we use max pooling for $\mathcal{F}_{\mathcal{P}}$ and mean aggregation for $\mathcal{F}_{\mathcal{G}}$. 
The number of sampled tokens $\mathcal{T}'$ is determined by the product of the total prompt length and the target retention ratio, with a $7{:}3$ recent-to-uniform sampling split.

\subsection{LongBench and RULER Results}{
\label{sec:accuracy_evaluation}
}

\begin{figure}
  \centering
  \begin{adjustbox}{max width=\textwidth}
    \includegraphics{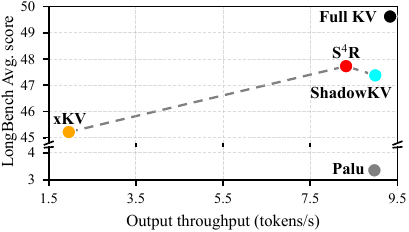}
  \end{adjustbox}
  
  \caption{
  Performance--speed trade-off on Qwen3-4B.
  The y-axis is broken to include the dominated Palu.
  }
  \label{fig:longbench_perf_speed}
\end{figure}

\begin{table*}
  \centering
  \small
  \setlength{\tabcolsep}{5pt}
  \begin{tabular}{lccccccccccccc}
    \toprule
    \multirow{2}{*}{Methods } & \multirow{2}{*}{$\eta$} & \multicolumn{3}{c}{Fewshot} & \multicolumn{3}{c}{Single-Doc QA} & \multicolumn{3}{c}{Multi-Doc QA} & \multirow{2}{*}{LCC} & \multirow{2}{*}{PRE}  & \multirow{2}{*}{Avg.} \\
    \cmidrule(lr){3-5}  \cmidrule(lr){6-8}  \cmidrule(lr){9-11}
    & & SSM & Trec & TQA  & MQA & NQA & QPR & 2Wiki & HQA & MSQ & \\
    \midrule
    \rowcolor{gray!20} Llama-3.2-1B & 1 & 28.73 & 63.50 & 80.35  & 41.09 & 18.29 & 13.02 & 29.74 & 31.73 & 14.65 & 23.76 & 2.50  & 31.58 \\
    Palu & \multirow{5}{*}{0.2} & 0.49 & 0.50 & 2.12  & 3.08 & 0.85 & 1.71 & 1.54 & 1.23 & 0.96 & 13.05 & 0.83  & 2.40 \\
    ShadowKV & & 19.97 & 45.00 & 73.55  & 30.67 & 14.96 & 11.62 & 23.33 & 29.08 & 9.79 & 19.46 & 5.50  & 25.72 \\
    xKV-2 & & 17.14 & 32.50 & 75.20  & 35.78 & 11.82 & 11.74 & 25.92 & 30.00 & 11.05 & 15.81 & 5.04  & 24.73 \\
    xKV-4 & & 20.58 & 35.50 & 77.08  & 36.46 & 14.87 & 11.81 & 25.78 & 28.92 & 12.74 & 16.33 & 6.01  & 26.01 \\
    \rowcolor{Salmon!40} S$^4$R & & 25.91 & 53.00 & 74.97 & 37.14 & 13.29 & 11.63 & 25.83 & 29.02 & 11.34 & 23.43 & 3.39 & 28.09 \\
    \midrule
    \rowcolor{gray!20} Qwen3-4B & 1 & 35.91 & 67.50 & 87.72  & 52.17 & 27.88 & 45.51 & 42.76 & 57.44 & 23.75 & 5.18 & 100.00 &  49.62 \\
    Palu & \multirow{5}{*}{0.2} & 0.89 & 2.50 & 4.43  & 3.83 & 0.60 & 1.64 & 0.92 & 0.75 & 0.83 & 19.82 & 0.64  & 3.35 \\
    ShadowKV & & 34.14 & 68.50 & 85.08  & 48.08 & 25.80 & 42.79 & 39.60 & 57.67 & 24.35 & 5.12 & 90.00  & 47.38 \\
    xKV-2 & & 32.88 & 41.00 & 88.06  & 51.02 & 22.80 & 41.58 & 38.95 & 53.80 & 19.21 & 4.92 & 89.75  & 44.00 \\ 
    xKV-4 & & 34.27 & 36.50 & 86.54  & 50.58 & 25.69 & 42.78 & 42.28 & 54.91 & 22.41 & 4.95 & 96.50  & 45.22 \\
    \rowcolor{Salmon!40} S$^4$R & & 34.38 & 72.50 & 89.31 & 50.92 & 23.81 & 40.85 & 39.93 & 53.90 & 18.84 & 5.05 & 96.17 & 47.79 \\
    \midrule
    \rowcolor{gray!20} Llama-3.1-8B & 1 & 43.96 & 73.50 & 92.29  & 56.34 & 30.51 & 45.59 & 50.02 & 58.22 & 31.75 & 24.76 & 100.00  & 55.18 \\
    Palu & \multirow{5}{*}{0.2} & 4.99 & 0.00 & 2.58  & 2.23 & 0.84 & 3.00 & 1.45 & 0.98 & 0.72 & 24.51 & 0.27  & 3.78 \\
    ShadowKV & & 45.10 & 68.00 & 92.34  & 35.65 & 29.84 & 24.81 & 49.24 & 58.13 & 31.02 & 25.08 & 92.00  & 50.11 \\
    xKV-2 & & 34.52 & 65.00 & 89.92  & 56.60 & 31.27 & 44.06 & 48.52 & 57.68 & 29.96 & 32.10 & 99.50  & 53.56 \\
    xKV-4 & & 35.16 & 63.50 & 91.00  & 56.94 & 32.27 & 45.96 & 50.21 & 57.99 & 30.61 & 32.51 & 100.00  & 54.19 \\
    \rowcolor{Salmon!40} S$^4$R & & 33.42 & 59.50 & 92.83 & 54.46 & 30.81 & 41.68 & 49.47 & 57.92 & 31.44 & 32.29 & 99.50  & 53.03 \\
    \midrule
    \rowcolor{gray!20} Qwen2.5-14B & 1 & 38.48 & 78.50 & 90.12 & 52.93 & 26.79 & 44.94 & 58.99 & 61.14 & 38.54 & 20.17 & 98.50 & 55.37 \\
    Palu & \multirow{5}{*}{0.3} & 20.76 & 0.64 & 15.29  & 22.32 & 3.42 & 8.44 & 10.63 & 9.23 & 5.51 & 16.59  & 28.33 & 12.83  \\
    ShadowKV & & 36.52 & 69.25 & 80.05  & 45.96 & 27.50 & 40.74 & 56.31 & 60.67 & 38.86 & 18.85  & 98.50 & 52.11 \\
    xKV-2 & & 37.17 & 77.50 & 91.02  & 51.50 & 27.08 & 45.06 & 57.02 & 59.50 & 38.51  & 19.41 & 99.00 & 54.80 \\
    xKV-4 & & 38.10 & 76.00 & 90.18  & 50.97 & 27.07 & 43.95 & 57.61 & 61.13 & 39.12 & 19.52 & 97.92  & 54.69\\
    \rowcolor{Salmon!40} S$^4$R & & 38.03 & 72.50 & 89.61  & 48.02 & 27.08 & 39.49 & 56.33 & 59.78 & 39.52  & 20.25 & 94.50 & 53.19 \\
    \bottomrule
  \end{tabular}
  \caption{
  Performance comparison of different models and methods on LongBench. 
  }
  \label{tab:exp_longbench}
\end{table*}

\textbf{LongBench}\quad Figure~\ref{fig:longbench_perf_speed} summarizes the main accuracy--efficiency trade-off on Qwen3-4B by combining the LongBench average scores in Table~\ref{tab:exp_longbench} with the output-throughput measurements in Table~\ref{tab:efficiency}.
S$^4$R lies on the favorable part of the trade-off curve: it achieves a LongBench average score of 47.73 (Table~\ref{tab:query_window_comparison}) under the latency-oriented efficiency setting, close to the default LongBench result of 47.79 in Table~\ref{tab:exp_longbench}, while reaching 8.32 output tokens/s in Table~\ref{tab:efficiency}.
In contrast, xKV has a lower Qwen3-4B LongBench average score in Table~\ref{tab:exp_longbench} and is much slower in Table~\ref{tab:efficiency}, with output throughput dropping to 1.96 tokens/s and 512-token latency increasing to 260.76s.
Palu and ShadowKV are faster, but Table~\ref{tab:exp_longbench} shows that Palu collapses under aggressive compression and ShadowKV remains below S$^4$R on Qwen3-4B accuracy.

Table~\ref{tab:exp_longbench} further expands this comparison across model families.
S$^4$R substantially outperforms Palu across all evaluated models, showing that prompt-aware subspace construction is important when the KV cache is compressed to 20--30\% of its original size.
Compared with ShadowKV and xKV, S$^4$R achieves the best average score on Llama-3.2-1B and Qwen3-4B, and remains competitive on larger models.
On Llama-3.1-8B and Qwen2.5-14B, xKV obtains slightly higher average scores, but this accuracy difference should be interpreted together with the efficiency gap.
Overall, Figure~\ref{fig:longbench_perf_speed}, Table~\ref{tab:exp_longbench}, and Table~\ref{tab:efficiency} show that S$^4$R preserves most of the LongBench accuracy of stronger prompt-dependent low-rank baselines while avoiding their large runtime overhead.

\noindent\textbf{RULER}\quad Table~\ref{tab:exp_ruler} reports the RULER results under a 2.5$\times$ compression ratio and a 128k context window.
In addition to low-rank baselines, we include two representative eviction-based methods, StreamingLLM (SLM) and SnapKV, to compare against token-retention strategies.
S$^4$R achieves an average score of 93.61, nearly matching the full KV cache result of 93.97 and slightly outperforming xKV at 93.47.
It also improves over Palu, ShadowKV, SnapKV, and SLM by 11.39, 4.00, 4.80, and 45.22 points on average, respectively.
The gains are especially clear on multi-key and multi-query retrieval tasks, where S$^4$R preserves perfect accuracy, suggesting that the sampled low-rank subspace and sparse reconstruction strategy retain the information needed for long-context retrieval.
The main remaining gap appears on FWE, where S$^4$R trails full KV and xKV, indicating that some tasks remain sensitive to the exact retained or reconstructed token set.

\begin{table}
  \centering
  \small
  \setlength{\tabcolsep}{1pt}
  \begin{tabular}{lccccccc}
    \toprule
    Method & MK1 & MK2 & MQ & MV & VT & FWE & Avg. \\
    \midrule
    \rowcolor{gray!20} Full KV & 100.00 & 100.00 & 100.00 & 97.50 & 83.00 & 83.33 & 93.97\\ 
    \midrule
    SLM &  55.00 & 30.00 & 35.00 & 35.00 & 47.00 & 88.33 & 48.39 \\
    SnapKV & 100.00 & 70.00 & 100.00 & 97.50 & 87.00 & 78.33 & 88.81\\
    Palu & 95.00 & 85.00 & 92.50 & 97.50 & 70.00 & 53.33 & 82.22 \\
    ShadowKV & 100.00 & 90.00 & 97.50 & 92.50 & 91.00 & 66.67 & 89.61 \\
    xKV & 100.00 & 100.00 & 97.50 & 100.00 & 80.00 & 83.33 & 93.47 \\
    \rowcolor{Salmon!40} S$^4$R & 100.00 & 100.00 & 100.00 & 100.00 & 85.00 & 76.67 & 93.61\\
    \bottomrule
  \end{tabular}
  \caption{
  Comparison of different methods on RULER using Qwen2.5-7B-Instruct-1M under a 2.5$\times$ compression ratio and a 128k context window. 
  }
  \label{tab:exp_ruler}
\end{table}

\begin{table*}
  \centering
  \small
  \setlength{\tabcolsep}{5pt}
  \begin{threeparttable}
  \begin{tabular}{lccccc}
    \toprule
    Method & TTFT (ms) $\downarrow$ & Latency$_{512}$ (s) $\downarrow$ & TPOT (ms/token) $\downarrow$ & Output Thr. (tokens/s) $\uparrow$ & Total Thr. (tokens/s) $\uparrow$ \\
    \midrule
    \rowcolor{gray!20} Full KV & 24737.54 & 54.89 & 59.01 & 9.33 & 2018.49 \\
    Palu & 24579.62 & 57.06 & 63.57 & 8.97 & 1941.67 \\
    ShadowKV\tnote{$\dagger$} & 36143.18 & 56.92 & 40.66 & 8.99 & 1946.43 \\
    xKV & 79897.02 & 260.76 & 353.95 & 1.96 & 424.89 \\
    \rowcolor{Salmon!40} S$^4$R\tnote{$\ddagger$} & 27650.96 & 61.54 & 66.19 & 8.32 & 1772.7 \\
    \bottomrule
  \end{tabular}
  \caption{
  Model-level efficiency comparison on Qwen3-4B using a 120k-token context sampled from GovReport.
  Latency$_{512}$ denotes the average end-to-end latency for generating 512 output tokens.
  All methods are evaluated without system-level, kernel-level, or operator-level optimizations.
  Appendix~\ref{sec:appendix_decoding_complexity} provides the corresponding theoretical efficiency estimates.
  }
  \label{tab:efficiency}
  \begin{tablenotes}[flushleft]
  \item[$\dagger$] For ShadowKV, we keep values on GPU rather than following its standard CPU-offloading implementation, which removes CPU--GPU value-loading communication overhead and therefore gives ShadowKV a more favorable efficiency estimate.
  \item[$\ddagger$]  For S$^4$R, this efficiency experiment uses a latency-oriented query window $\mathcal{W}=4$. Appendix~\ref{sec:appendix_effect_query_window} shows that this has negligible impact on Qwen3-4B accuracy, with an average score of 47.73 compared with 47.79 for $\mathcal{W}=32$. 
  \end{tablenotes}
  \end{threeparttable}
\end{table*}

\subsection{Efficiency Evaluation}

We report model-level generation efficiency using TTFT, average latency for generating 512 output tokens, time per output token (TPOT), output throughput, and total throughput.

Table~\ref{tab:efficiency} shows that S$^4$R introduces moderate overhead compared with the full KV baseline and fixed/offline-style methods, mainly because it performs latent-space scoring and sparse reconstruction during decoding.
Nevertheless, the overhead is small in this setting: S$^4$R reaches a 512-token latency of 61.54s, close to Full KV, Palu, and ShadowKV at 54.89s, 57.06s, and 56.92s, respectively.
More importantly, S$^4$R is substantially more efficient than xKV: it reduces TTFT from 79.90s to 27.65s and the 512-token generation latency from 260.76s to 61.54s, while improving output throughput from 1.96 to 8.32 tokens/s.
This supports the intended trade-off of S$^4$R: compared with more expensive prompt-dependent low-rank approaches, it retains prompt adaptivity while avoiding the cost of decomposing or processing the full prompt during inference.
Compared with Palu and ShadowKV, S$^4$R is slightly slower in this unoptimized implementation, but the LongBench and RULER results show that the additional runtime is exchanged for stronger accuracy under aggressive KV compression.

\subsection{Ablation Study}

We ablate three components that are central to S$^4$R in the main text.
Unless otherwise specified, ablation tables report grouped LongBench scores: Fewshot, SDQA, and MDQA denote the average score within each task group, while Avg. is computed over the same 11 tasks as Table~\ref{tab:exp_longbench}.
Additional ablations are provided in Appendix~\ref{sec:appendix_more_ablations}.

\subsubsection{Importance of Sink Token}
\label{sec:sink_token}

S$^4$R treats the first few tokens as attention sinks and keeps their key/value states in full precision.
Unlike ordinary non-sink tokens, compressing these positions into a low-rank coefficient space may introduce errors that are repeatedly exposed to many later queries.
We therefore isolate the effect of sink-token preservation by comparing the default setting, which stores the first $s=4$ sink tokens in full precision, with a variant that applies the same low-rank compression pipeline to all tokens.

\begin{table}
  \centering
  \small
  \setlength{\tabcolsep}{3pt}
  \begin{tabular}{ccccccc}
    \toprule
    Sink & Fewshot & SDQA & MDQA & LCC & PRE & Avg. \\
    \midrule
    \multicolumn{7}{c}{Llama-3.2-1B}\\
    \rowcolor{Salmon!40} w/ & 51.29 & 20.69 & 22.06 & 23.43 & 3.39 & 28.09 \\
    w/o & 42.21 & 18.82 & 20.86 & 20.56 & 4.36 & 24.60 \\
    \midrule
    \multicolumn{7}{c}{Qwen3-4B}\\
    \rowcolor{Salmon!40} w/ & 65.40 & 38.53 & 37.56 & 5.05 & 96.17 & 47.79 \\
    w/o & 62.99 & 37.00 & 38.11 & 5.04 & 82.00 & 45.58 \\
    \midrule
    \multicolumn{7}{c}{Llama-3.1-8B}\\
    \rowcolor{Salmon!40} w/ & 61.92 & 42.32 & 46.28 & 32.29 & 99.50 & 53.03 \\
    w/o & 60.51 & 41.92 & 45.71 & 31.59 & 99.50 & 52.32 \\
    \midrule
    \multicolumn{7}{c}{Qwen2.5-14B}\\
    \rowcolor{Salmon!40} w/ & 66.71 & 38.20 & 51.88 & 20.25 & 94.50 & 53.19 \\
    w/o & 64.88 & 37.29 & 50.56 & 19.80 & 94.50 & 52.04 \\
    \bottomrule
  \end{tabular}
  \caption{Effect of retaining sink tokens in full precision on grouped LongBench performance. ``w/'' keeps the first $s=4$ sink tokens uncompressed, whereas ``w/o'' applies the low-rank compression pipeline to all tokens.}
  \label{tab:ablation_sink_token}
\end{table}

Table~\ref{tab:ablation_sink_token} reports the grouped LongBench results.
Retaining sink tokens in full precision improves the average score across all four models.
The gain is largest on smaller models: the average score increases from 24.60 to 28.09 on Llama-3.2-1B and from 45.58 to 47.79 on Qwen3-4B.
The improvement remains consistent on larger models, with gains of 0.71 points on Llama-3.1-8B and 1.15 points on Qwen2.5-14B.
These results indicate that the first few sink tokens provide a stable global context that is difficult to recover once compressed into the same low-rank coefficient space as ordinary tokens.
The task-level effects are also informative: Llama-3.2-1B benefits most on Fewshot and LCC, while Qwen3-4B shows a large improvement on passage retrieval, where PRE increases from 82.00 to 96.17.
Since the sink cache only contributes $2s\mathcal{D}'$ elements per layer in Eq.~(\ref{eq:memory_ratio}), its memory overhead is negligible for long contexts.
% We therefore keep full-precision sink-token preservation as the default because it provides a consistent accuracy gain at minimal memory cost.

\subsubsection{Effect of Token-wise Normalization}
\label{sec:token_wise_norm}

We ablate the token-wise normalization step before SVD in Table~\ref{tab:ablation_normalization}.
On Llama-3.1-8B, removing normalization leads to a slightly higher average score, with 53.32 compared to 53.03 for the default normalized variant.
However, the effect is model-dependent: on Qwen2.5-14B, removing normalization causes a substantial drop from 53.19 to 46.27.
The degradation is especially pronounced on single-document QA, multi-document QA, and passage retrieval, indicating that unnormalized high-magnitude KV states can dominate the SVD basis and produce less transferable low-rank subspaces.
We therefore keep token-wise normalization as the default because it improves robustness across model families, even though it is not uniformly optimal on every model.

\begin{table}[t]
  \centering
  \small
  \setlength{\tabcolsep}{3pt}
  \begin{tabular}{ccccccc}
    \toprule
    Norm. & Fewshot & SDQA & MDQA & LCC & PRE & Avg. \\
    \midrule
    \multicolumn{7}{c}{Llama-3.1-8B}\\
    \rowcolor{Salmon!40} w/ & 61.92 & 42.32 & 46.28 & 32.29 & 99.50 & 53.03 \\
    w/o & 62.78 & 42.41 & 46.13 & 32.54 & 100.00 & 53.32 \\
    \midrule
    \multicolumn{7}{c}{Qwen2.5-14B}\\
    \rowcolor{Salmon!40} w/ & 66.71 & 38.20 & 51.88 & 20.25 & 94.50 & 53.19 \\
    w/o & 63.37 & 31.43 & 43.01 & 18.77 & 76.79 & 46.27 \\
    \bottomrule
  \end{tabular}
  \caption{
  Effect of token-wise normalization before SVD. Norm. indicates whether sampled KV states are normalized before constructing the subspaces.
  }
  \label{tab:ablation_normalization}
\end{table}

\subsubsection{Pooling and Aggregation}
\label{sec:pooling_aggregation}

We study how the pooling function $\mathcal{F}_{\mathcal{P}}$ over neighboring token scores and the aggregation function $\mathcal{F}_{\mathcal{G}}$ affect the final sparse reconstruction set.
Table~\ref{tab:ablation_pooling_aggregation} compares max and average pooling along the sequence dimension, combined with mean or max aggregation across group dimensions.
Across all four model settings, the average score varies only mildly across these choices: the largest gap among the four variants is 0.38 for Llama-3.2-1B, 0.23 for Qwen3-4B, 0.67 for Llama-3.1-8B, and 0.18 for Qwen2.5-14B.
This suggests that S$^4$R is not highly sensitive to the exact smoothing and group-aggregation rule used for latent-space relevance scores.

\begin{table}[t]
  \centering
  \small
  \setlength{\tabcolsep}{4.5pt}
  \begin{tabular}{cccccc}
    \toprule
    $\mathcal{F}_{\mathcal{P}}$ & $\mathcal{F}_{\mathcal{G}}$ & Fewshot & SDQA & MDQA & Group Avg. \\
    \midrule
    \multicolumn{6}{c}{Llama-3.2-1B} \\
    \rowcolor{Salmon!40} $\mathrm{max}$ & $\mathrm{mean}$ & 51.29 & 20.69 & 22.06 & 31.35 \\
    $\mathrm{max}$ & $\mathrm{max}$ & 51.27 & 19.77 & 22.67 & 31.24 \\
    $\mathrm{avg}$ & $\mathrm{mean}$ & 52.24 & 20.03 & 22.52 & 31.60 \\
    $\mathrm{avg}$ & $\mathrm{max}$ & 51.92 & 20.25 & 22.69 & 31.62 \\
    \midrule
    \multicolumn{6}{c}{Qwen3-4B} \\
    \rowcolor{Salmon!40} $\mathrm{max}$ & $\mathrm{mean}$ & 65.40 & 38.53 & 37.56 & 47.16 \\
    $\mathrm{max}$ & $\mathrm{max}$ & 65.36 & 37.89 & 37.56 & 46.94 \\
    $\mathrm{avg}$ & $\mathrm{mean}$ & 64.76 & 38.30 & 37.74 & 46.93 \\
    $\mathrm{avg}$ & $\mathrm{max}$ & 65.31 & 38.59 & 37.38 & 47.09 \\
    \midrule
    \multicolumn{6}{c}{Llama-3.1-8B} \\
    \rowcolor{Salmon!40} $\mathrm{max}$ & $\mathrm{mean}$ & 61.92 & 42.32 & 46.28 & 50.17 \\
    $\mathrm{max}$ & $\mathrm{max}$ & 61.70 & 42.47 & 46.66 & 50.28 \\
    $\mathrm{avg}$ & $\mathrm{mean}$ & 60.69 & 42.49 & 46.37 & 49.85 \\
    $\mathrm{avg}$ & $\mathrm{max}$ & 60.17 & 42.20 & 46.47 & 49.61 \\
    \midrule
    \multicolumn{6}{c}{Qwen2.5-14B} \\
    \rowcolor{Salmon!40} $\mathrm{max}$ & $\mathrm{mean}$ & 66.71 & 38.20 & 51.88 & 52.26 \\
    $\mathrm{max}$ & $\mathrm{max}$ & 66.66 & 37.98 & 52.02 & 52.22 \\
    $\mathrm{avg}$ & $\mathrm{mean}$ & 66.77 & 37.50 & 51.98 & 52.08 \\
    $\mathrm{avg}$ & $\mathrm{max}$ & 66.78 & 37.79 & 52.01 & 52.19 \\
    \bottomrule
  \end{tabular}
  \caption{
  Effect of pooling and aggregation choices for latent-space relevance estimation.
  }
  \label{tab:ablation_pooling_aggregation}
\end{table}

\section{Conclusion}

We presented S$^4$R, a prompt-aware low-rank KV cache compression method that combines selective subspace construction with sparse reconstruction.
Instead of relying on external calibration data or decomposing the full prompt, S$^4$R constructs key/value bases from sampled non-sink prompt tokens, stores non-sink KV states as latent coefficients, preserves sink tokens in full precision, and reconstructs only the local window plus globally important positions during decoding.
This design targets the main trade-off in low-rank KV compression: fixed methods are efficient but less adaptive, while prompt-dependent methods are adaptive but can be costly on long prompts.

Experiments on LongBench and RULER show that S$^4$R maintains strong long-context accuracy under aggressive KV compression.
On LongBench, it outperforms fixed low-rank baselines and remains competitive with more expensive prompt-dependent methods; on RULER, it achieves 93.61 average score under 2.5$\times$ compression, nearly matching full KV and slightly surpassing xKV.
Efficiency results further show that S$^4$R substantially reduces the overhead of xKV in a 120k-token setting, lowering TTFT from 79.90s to 27.65s and 512-token latency from 260.76s to 61.54s.
Together, these results indicate that sampled-token SVD and latent-space sparse reconstruction provide a practical path toward adaptive KV cache compression for long-context LLM inference.

\section*{Limitations}
Our implementation lacks system- or kernel-level optimization, so reported efficiency reflects algorithmic overhead rather than a fully optimized deployment.
Experiments are limited to Llama and Qwen on LongBench and RULER, which emphasize long-input understanding with relatively short outputs; behavior under very long generation, other architectures, and broader workloads remains to be validated.
S$^4$R also depends on hyperparameters and a fixed token-sampling heuristic, and tasks with scattered evidence (e.g., FWE) can still degrade when latent scoring misses relevant positions.

\section*{Acknowledgments}
This work was supported by the HPC platform of ShanghaiTech University, and the Core Facility Platform of Computer Science and Communication, SIST, ShanghaiTech University.


\bibliography{custom}

@inproceedings{adakv2025,
 author = {Feng, Yuan and Lv, Junlin and Cao, Yukun and Xie, Xike and Zhou, S. Kevin},
 booktitle = {Advances in Neural Information Processing Systems},
 doi = {10.52202/085713-3773},
 editor = {D. Belgrave and C. Zhang and H. Lin and R. Pascanu and P. Koniusz and M. Ghassemi and N. Chen},
 pages = {113152--113188},
 publisher = {Curran Associates, Inc.},
 title = {{Ada-KV: Optimizing KV Cache Eviction by Adaptive Budget Allocation for Efficient LLM Inference}},
 url = {https://proceedings.neurips.cc/paper_files/paper/2025/file/a40ff56daab9f4808b1e18350c8a11ce-Paper-Conference.pdf},
 volume = {38, Main Conference},
 year = {2025}
}

@inproceedings{cam,
  title = 	 {{CaM: Cache Merging for Memory-efficient LLMs Inference}},
  author =       {Zhang, Yuxin and Du, Yuxuan and Luo, Gen and Zhong, Yunshan and Zhang, Zhenyu and Liu, Shiwei and Ji, Rongrong},
  booktitle = 	 {Proceedings of the 41st International Conference on Machine Learning},
  pages = 	 {58840--58850},
  year = 	 {2024},
  editor = 	 {Salakhutdinov, Ruslan and Kolter, Zico and Heller, Katherine and Weller, Adrian and Oliver, Nuria and Scarlett, Jonathan and Berkenkamp, Felix},
  volume = 	 {235},
  series = 	 {Proceedings of Machine Learning Research},
  month = 	 {21--27 Jul},
  publisher =    {PMLR},
  url = 	 {https://proceedings.mlr.press/v235/zhang24n.html},
  address = {Vienna, Austria},
}

@misc{commonkv2025,
      title={{CommonKV: Compressing KV Cache with Cross-layer Parameter Sharing}}, 
      author={Yixuan Wang and Haoyu Qiao and Lujun Li and Qingfu Zhu and Wanxiang Che},
      year={2025},
      eprint={2508.16134},
      archivePrefix={arXiv},
      primaryClass={cs.LG},
      url={https://arxiv.org/abs/2508.16134}, 
}

@inproceedings{duoattention2025,
 author = {Xiao, Guangxuan and Tang, Jiaming and Zuo, Jingwei and Guo, Junxian and Yang, Shang and Tang, Haotian and Fu, Yao and Han, Song},
 booktitle = {International Conference on Learning Representations},
 editor = {Y. Yue and A. Garg and N. Peng and F. Sha and R. Yu},
 pages = {37228--37253},
 title = {{DuoAttention: Efficient Long-Context LLM Inference with Retrieval and Streaming Heads}},
 url = {https://proceedings.iclr.cc/paper_files/paper/2025/file/5c1ddd2e59df46fd2aa85c833b1b36ed-Paper-Conference.pdf},
 volume = {2025},
 address = {Singapore},
 year = {2025}
}

@misc{dynakv2025,
      title={{DynaKV: Enabling Accurate and Efficient Long-Sequence LLM Decoding on Smartphones}}, 
      author={Tuowei Wang and Minxing Huang and Fengzu Li and Ligeng Chen and Jinrui Zhang and Ju Ren},
      year={2025},
      eprint={2511.07427},
      archivePrefix={arXiv},
      primaryClass={cs.DC},
      url={https://arxiv.org/abs/2511.07427}, 
}

@misc{effectively2025,
      title={{Effectively Compress KV Heads for LLM}},
      author={Hao Yu and Zelan Yang and Shen Li and Yong Li and Jianxin Wu},
      year={2024},
      eprint={2406.07056},
      archivePrefix={arXiv},
      primaryClass={cs.CL},
      url={https://arxiv.org/abs/2406.07056},
}

@inproceedings{efficient2023,
author = {Kwon, Woosuk and Li, Zhuohan and Zhuang, Siyuan and Sheng, Ying and Zheng, Lianmin and Yu, Cody Hao and Gonzalez, Joseph and Zhang, Hao and Stoica, Ion},
title = {{Efficient Memory Management for Large Language Model Serving with PagedAttention}},
year = {2023},
isbn = {9798400702297},
publisher = {Association for Computing Machinery},
address = {New York, NY, USA},
url = {https://doi.org/10.1145/3600006.3613165},
doi = {10.1145/3600006.3613165},
booktitle = {Proceedings of the 29th Symposium on Operating Systems Principles},
pages = {611–626},
numpages = {16},
location = {Koblenz, Germany},
series = {SOSP '23}
}

@inproceedings{MLSYS2023_c4be71ab,
  address = {Miami Beach, Florida, USA},
 author = {Pope, Reiner and Douglas, Sholto and Chowdhery, Aakanksha and Devlin, Jacob and Bradbury, James and Heek, Jonathan and Xiao, Kefan and Agrawal, Shivani and Dean, Jeff},
 booktitle = {Proceedings of Machine Learning and Systems},
 editor = {D. Song and M. Carbin and T. Chen},
 pages = {606--624},
 publisher = {Curan},
 title = {{Efficiently Scaling Transformer Inference}},
 url = {https://proceedings.mlsys.org/paper_files/paper/2023/file/c4be71ab8d24cdfb45e3d06dbfca2780-Paper-mlsys2023.pdf},
 volume = {5},
 year = {2023}
}

@inproceedings{efficient2024,
 author = {Xiao, Guangxuan and Tian, Yuandong and Chen, Beidi and Han, Song and Lewis, Mike },
 booktitle = {International Conference on Learning Representations},
 editor = {B. Kim and Y. Yue and S. Chaudhuri and K. Fragkiadaki and M. Khan and Y. Sun},
 pages = {21875--21895},
 title = {{Efficient Streaming Language Models with Attention Sinks}},
 url = {https://proceedings.iclr.cc/paper_files/paper/2024/file/5e5fd18f863cbe6d8ae392a93fd271c9-Paper-Conference.pdf},
 volume = {2024},
 address = {Vienna, Austria},
 year = {2024}
}

@inproceedings{gear2024,
  title = 	 {{GEAR: An Efficient Error Reduction Framework for KV Cache Compression in LLM Inference}},
  author =       {Kang, Hao and Zhang, Qingru and Kundu, Souvik and Jeong, Geonhwa and Liu, Zaoxing and Krishna, Tushar and Zhao, Tuo},
  booktitle = 	 {Proceedings of The 4th NeurIPS Efficient Natural Language and Speech Processing Workshop},
  pages = 	 {305--321},
  year = 	 {2024},
  editor = 	 {Rezagholizadeh, Mehdi and Passban, Peyman and Samiee, Soheila and Partovi Nia, Vahid and Cheng, Yu and Deng, Yue and Liu, Qun and Chen, Boxing},
  volume = 	 {262},
  series = 	 {Proceedings of Machine Learning Research},
  month = 	 {14 Dec},
  publisher =    {PMLR},
  url = 	 {https://proceedings.mlr.press/v262/kang24a.html},
  address = {Vancouver, Canada}
}

@inproceedings{h2o2023,
 author = {Zhang, Zhenyu and Sheng, Ying and Zhou, Tianyi and Chen, Tianlong and Zheng, Lianmin and Cai, Ruisi and Song, Zhao and Tian, Yuandong and R\'{e}, Christopher and Barrett, Clark and Wang, Zhangyang "Atlas" and Chen, Beidi},
 booktitle = {Advances in Neural Information Processing Systems},
 doi = {10.52202/075280-1506},
 editor = {A. Oh and T. Naumann and A. Globerson and K. Saenko and M. Hardt and S. Levine},
 pages = {34661--34710},
 publisher = {Curran Associates, Inc.},
 title = {{H2O: Heavy-Hitter Oracle for Efficient Generative Inference of Large Language Models}},
 url = {https://proceedings.neurips.cc/paper_files/paper/2023/file/6ceefa7b15572587b78ecfcebb2827f8-Paper-Conference.pdf},
 volume = {36},
 address = {New Orleans, LA, USA},
 year = {2023}
}

@inproceedings{kivi2023,
  title = 	 {{KIVI: A Tuning-Free Asymmetric 2bit Quantization for KV Cache}},
  author =       {Liu, Zirui and Yuan, Jiayi and Jin, Hongye and Zhong, Shaochen and Xu, Zhaozhuo and Braverman, Vladimir and Chen, Beidi and Hu, Xia},
  booktitle = 	 {Proceedings of the 41st International Conference on Machine Learning},
  pages = 	 {32332--32344},
  year = 	 {2024},
  editor = 	 {Salakhutdinov, Ruslan and Kolter, Zico and Heller, Katherine and Weller, Adrian and Oliver, Nuria and Scarlett, Jonathan and Berkenkamp, Felix},
  volume = 	 {235},
  series = 	 {Proceedings of Machine Learning Research},
  month = 	 {21--27 Jul},
  publisher =    {PMLR},
  url = 	 {https://proceedings.mlr.press/v235/liu24bz.html},
  address = {Vienna, Austria},
}

@misc{kvcompress2024,
      title={{KV-Compress: Paged KV-Cache Compression with Variable Compression Rates per Attention Head}}, 
      author={Isaac Rehg},
      year={2024},
      eprint={2410.00161},
      archivePrefix={arXiv},
      primaryClass={cs.CL},
      url={https://arxiv.org/abs/2410.00161}, 
}

@inproceedings{kvquant,
  address = {Vancouver, Canada},
   author = {Hooper, Coleman and Kim, Sehoon and Mohammadzadeh, Hiva and Mahoney, Michael W. and Shao, Yakun Sophia and Keutzer, Kurt and Gholami, Amir},
 booktitle = {Advances in Neural Information Processing Systems},
 doi = {10.52202/079017-0040},
 editor = {A. Globerson and L. Mackey and D. Belgrave and A. Fan and U. Paquet and J. Tomczak and C. Zhang},
 pages = {1270--1303},
 publisher = {Curran Associates, Inc.},
 title = {{KVQuant: Towards 10 Million Context Length LLM Inference with KV Cache Quantization}},
 url = {https://proceedings.neurips.cc/paper_files/paper/2024/file/028fcbcf85435d39a40c4d61b42c99a4-Paper-Conference.pdf},
 volume = {37},
 year = {2024}
}

@inproceedings{kvzip2025,
 author = {Kim, Jang-Hyun and Kim, Jinuk and Kwon, Sangwoo and Lee, Jae W. and Yun, Sangdoo and Song, Hyun Oh},
 booktitle = {Advances in Neural Information Processing Systems},
 doi = {10.52202/085713-5585},
 editor = {D. Belgrave and C. Zhang and H. Lin and R. Pascanu and P. Koniusz and M. Ghassemi and N. Chen},
 pages = {167563--167591},
 publisher = {Curran Associates, Inc.},
 title = {{KVzip: Query-Agnostic KV Cache Compression with Context Reconstruction}},
 url = {https://proceedings.neurips.cc/paper_files/paper/2025/file/f4eaa4b8f2d08edb3f0af990d56134ea-Paper-Conference.pdf},
 volume = {38, Main Conference},
 address = {San Diego, CA, USA},
 year = {2025}
}

@inproceedings{loki2024,
 author = {Singhania, Prajwal and Singh, Siddharth and He, Shwai and Feizi, Soheil and Bhatele, Abhinav},
 booktitle = {Advances in Neural Information Processing Systems},
 doi = {10.52202/079017-0532},
 editor = {A. Globerson and L. Mackey and D. Belgrave and A. Fan and U. Paquet and J. Tomczak and C. Zhang},
 pages = {16692--16723},
 publisher = {Curran Associates, Inc.},
 title = {{Loki: Low-rank Keys for Efficient Sparse Attention}},
 url = {https://proceedings.neurips.cc/paper_files/paper/2024/file/1e027da6bec9ceb2ec37951ceeccae93-Paper-Conference.pdf},
 volume = {37},
 year = {2024},
 address = {Vancouver, Canada}
}

@misc{lorc2024,
      title={{LoRC: Low-Rank Compression for LLMs KV Cache with a Progressive Compression Strategy}}, 
      author={Rongzhi Zhang and Kuang Wang and Liyuan Liu and Shuohang Wang and Hao Cheng and Chao Zhang and Yelong Shen},
      year={2024},
      eprint={2410.03111},
      archivePrefix={arXiv},
      primaryClass={cs.LG},
      url={https://arxiv.org/abs/2410.03111}, 
}

@misc{model2024,
      title={{Model Tells You Where to Merge: Adaptive KV Cache Merging for LLMs on Long-Context Tasks}}, 
      author={Zheng Wang and Boxiao Jin and Zhongzhi Yu and Minjia Zhang},
      year={2024},
      eprint={2407.08454},
      archivePrefix={arXiv},
      primaryClass={cs.CL},
      url={https://arxiv.org/abs/2407.08454}, 
}

@inproceedings{nacl2024,
  title = {{NACL: A General and Effective KV Cache Eviction Framework for LLM at Inference Time}},
  shorttitle = {{{NACL}}},
  booktitle = {Proceedings of the 62nd {{Annual Meeting}} of the {{Association}} for {{Computational Linguistics}} ({{Volume}} 1: {{Long Papers}})},
  date = {2024-08},
  pages = {7913--7926},
  publisher = {Association for Computational Linguistics},
  address = {Bangkok, Thailand},
  doi = {10.18653/v1/2024.acl-long.428},
  url = {https://aclanthology.org/2024.acl-long.428/},
  urldate = {2025-08-12},
  eventtitle = {{{ACL}} 2024},
  year = {2024},
  author = {Chen, Yilong and Wang, Guoxia and Shang, Junyuan and Cui, Shiyao and Zhang, Zhenyu and Liu, Tingwen and Wang, Shuohuan and Sun, Yu and Yu, Dianhai and Wu, Hua}
}

@inproceedings{palu2025,
 author = {Chang, Chi-Chih and Lin, Wei-Cheng and Lin, Chien-Yu and Chen, Chong-Yan and Hu, Yu-Fang and Wang, Pei-Shuo and Huang, Ning-Chi and Ceze, Luis and Abdelfattah, Mohamed and Wu, Kai-Chiang},
 booktitle = {International Conference on Learning Representations},
 editor = {Y. Yue and A. Garg and N. Peng and F. Sha and R. Yu},
 pages = {50222--50249},
 title = {{Palu: KV-Cache Compression with Low-Rank Projection}},
 url = {https://proceedings.iclr.cc/paper_files/paper/2025/file/7da6e0e00702c60607a6ae05c802ef85-Paper-Conference.pdf},
 volume = {2025},
 address = {Singapore},
 year = {2025}
}

@inproceedings{pyramidkv2025,
title={{PyramidKV: Dynamic KV Cache Compression based on Pyramidal Information Funneling}},
author={Zefan Cai and Yichi Zhang and Bofei Gao and Yuliang Liu and Yucheng Li and Tianyu Liu and Keming Lu and Wayne Xiong and Yue Dong and Junjie Hu and Wen Xiao},
booktitle={Second Conference on Language Modeling},
year={2025},
url={https://openreview.net/forum?id=ayi7qezU87},
publisher={OpenReview.net},
address = {Montreal, Canada}
}

@inproceedings{quest2024,
  title = 	 {{QUEST: Query-Aware Sparsity for Efficient Long-Context LLM Inference}},
  author =       {Tang, Jiaming and Zhao, Yilong and Zhu, Kan and Xiao, Guangxuan and Kasikci, Baris and Han, Song},
  booktitle = 	 {Proceedings of the 41st International Conference on Machine Learning},
  pages = 	 {47901--47911},
  year = 	 {2024},
  editor = 	 {Salakhutdinov, Ruslan and Kolter, Zico and Heller, Katherine and Weller, Adrian and Oliver, Nuria and Scarlett, Jonathan and Berkenkamp, Felix},
  volume = 	 {235},
  series = 	 {Proceedings of Machine Learning Research},
  month = 	 {21--27 Jul},
  publisher =    {PMLR},
  url = 	 {https://proceedings.mlr.press/v235/tang24l.html},
  address = {Vienna, Austria},
}

@misc{recalkv2025,
      title={{ReCalKV: Low-Rank KV Cache Compression via Head Reordering and Offline Calibration}}, 
      author={Xianglong Yan and Zhiteng Li and Tianao Zhang and Haotong Qin and Linghe Kong and Yulun Zhang and Xiaokang Yang},
      year={2025},
      eprint={2505.24357},
      archivePrefix={arXiv},
      primaryClass={cs.LG},
      url={https://arxiv.org/abs/2505.24357}, 
}

@inproceedings{sals,
 author = {Mu, Junlin and Huang, Hantao and Zhang, Jihang and Yu, Minghui and Wang, Tao and Li, Yidong},
 booktitle = {Advances in Neural Information Processing Systems},
 doi = {10.52202/085713-0013},
 editor = {D. Belgrave and C. Zhang and H. Lin and R. Pascanu and P. Koniusz and M. Ghassemi and N. Chen},
 pages = {384--403},
 publisher = {Curran Associates, Inc.},
 title = {{SALS: Sparse Attention in Latent Space for KV Cache Compression}},
 url = {https://proceedings.neurips.cc/paper_files/paper/2025/file/00a0ebcad584c59dbc439c2af8793638-Paper-Conference.pdf},
 volume = {38, Main Conference},
 address = {San Diego, CA, USA},
 year = {2025}
}

@inproceedings{scissorhands2023,
 author = {Liu, Zichang and Desai, Aditya and Liao, Fangshuo and Wang, Weitao and Xie, Victor and Xu, Zhaozhuo and Kyrillidis, Anastasios and Shrivastava, Anshumali},
 booktitle = {Advances in Neural Information Processing Systems},
 doi = {10.52202/075280-2279},
 editor = {A. Oh and T. Naumann and A. Globerson and K. Saenko and M. Hardt and S. Levine},
 pages = {52342--52364},
 publisher = {Curran Associates, Inc.},
 title = {{Scissorhands: Exploiting the Persistence of Importance Hypothesis for LLM KV Cache Compression at Test Time}},
 url = {https://proceedings.neurips.cc/paper_files/paper/2023/file/a452a7c6c463e4ae8fbdc614c6e983e6-Paper-Conference.pdf},
 volume = {36},
 address = {New Orleans, LA, USA},
 year = {2023}
}

@inproceedings{shadowkv2025,
  title = 	 {{ShadowKV: KV Cache in Shadows for High-Throughput Long-Context LLM Inference}},
  author =       {Sun, Hanshi and Chang, Li-Wen and Bao, Wenlei and Zheng, Size and Zheng, Ningxin and Liu, Xin and Dong, Harry and Chi, Yuejie and Chen, Beidi},
  booktitle = 	 {Proceedings of the 42nd International Conference on Machine Learning},
  pages = 	 {57355--57373},
  year = 	 {2025},
  editor = 	 {Singh, Aarti and Fazel, Maryam and Hsu, Daniel and Lacoste-Julien, Simon and Berkenkamp, Felix and Maharaj, Tegan and Wagstaff, Kiri and Zhu, Jerry},
  volume = 	 {267},
  series = 	 {Proceedings of Machine Learning Research},
  month = 	 {13--19 Jul},
  publisher =    {PMLR},
  url = 	 {https://proceedings.mlr.press/v267/sun25b.html},
  address = {Vancouver, Canada},
}

@inproceedings{snapkv2024,
 author = {Li, Yuhong and Huang, Yingbing and Yang, Bowen and Venkitesh, Bharat and Locatelli, Acyr and Ye, Hanchen and Cai, Tianle and Lewis, Patrick and Chen, Deming},
 booktitle = {Advances in Neural Information Processing Systems},
 doi = {10.52202/079017-0722},
 editor = {A. Globerson and L. Mackey and D. Belgrave and A. Fan and U. Paquet and J. Tomczak and C. Zhang},
 pages = {22947--22970},
 publisher = {Curran Associates, Inc.},
 title = {{SnapKV: LLM Knows What You are Looking for Before Generation}},
 url = {https://proceedings.neurips.cc/paper_files/paper/2024/file/28ab418242603e0f7323e54185d19bde-Paper-Conference.pdf},
 volume = {37},
 year = {2024},
  address = {Vancouver, Canada},
}

@inproceedings{specache2025,
  title = 	 {{SpeCache: Speculative Key-Value Caching for Efficient Generation of LLMs}},
  author =       {Jie, Shibo and Tang, Yehui and Han, Kai and Deng, Zhi-Hong and Han, Jing},
  booktitle = 	 {Proceedings of the 42nd International Conference on Machine Learning},
  pages = 	 {27917--27928},
  year = 	 {2025},
  editor = 	 {Singh, Aarti and Fazel, Maryam and Hsu, Daniel and Lacoste-Julien, Simon and Berkenkamp, Felix and Maharaj, Tegan and Wagstaff, Kiri and Zhu, Jerry},
  volume = 	 {267},
  series = 	 {Proceedings of Machine Learning Research},
  month = 	 {13--19 Jul},
  publisher =    {PMLR},
  url = 	 {https://proceedings.mlr.press/v267/jie25a.html},
  address = {Vancouver, Canada}
}

@article{survey2025b,
  title={{A Survey on Large Language Model Acceleration based on KV Cache Management}},
  author={Haoyang Li and Yiming Li and Anxin Tian and Tianhao Tang and Zhanchao Xu and Xuejia Chen and Nicole HU and Wei Dong and Li Qing and Lei Chen},
  journal={Transactions on Machine Learning Research},
  issn={2835-8856},
  year={2025},
  url={https://openreview.net/forum?id=z3JZzu9EA3},
}

@inproceedings{xkv2025,
title={{xKV: Cross-Layer KV-Cache Compression via Aligned Singular Vector Extraction}},
author={Chi-Chih Chang and Wei-Cheng Lin and Chien-Yu Lin and Hung-Yueh Chiang and Yash Akhauri and Xilai Dai and Huiqiang Jiang and Yucheng Li and Luis Ceze and Kai-Chiang Wu and Mohamed S. Abdelfattah},
booktitle={Forty-third International Conference on Machine Learning},
year={2026},
url={https://openreview.net/forum?id=XDb69lykA8},
address = {Seoul, South Korea}
}

@misc{gemini2024,
title={{Gemini 1.5: Unlocking multimodal understanding across millions of tokens of context}}, 
      author={Gemini Team and Petko Georgiev and Ving Ian Lei and Ryan Burnell and Libin Bai and Anmol Gulati and Garrett Tanzer and Damien Vincent and Zhufeng Pan and Shibo Wang and Soroosh Mariooryad and Yifan Ding and Xinyang Geng and Fred Alcober and Roy Frostig and Mark Omernick and Lexi Walker and Cosmin Paduraru and Christina Sorokin and Andrea Tacchetti and Colin Gaffney and Samira Daruki and Olcan Sercinoglu and Zach Gleicher and Juliette Love and Paul Voigtlaender and Rohan Jain and Gabriela Surita and Kareem Mohamed and Rory Blevins and Junwhan Ahn and Tao Zhu and Kornraphop Kawintiranon and Orhan Firat and Yiming Gu and Yujing Zhang and Matthew Rahtz and Manaal Faruqui and Natalie Clay and Justin Gilmer and JD Co-Reyes and Ivo Penchev and Rui Zhu and Nobuyuki Morioka and Kevin Hui and Krishna Haridasan and Victor Campos and Mahdis Mahdieh and Mandy Guo and Samer Hassan and Kevin Kilgour and Arpi Vezer and Heng-Tze Cheng and Raoul de Liedekerke and Siddharth Goyal and Paul Barham and DJ Strouse and Seb Noury and Jonas Adler and Mukund Sundararajan and Sharad Vikram and Dmitry Lepikhin and Michela Paganini and Xavier Garcia and Fan Yang and Dasha Valter and Maja Trebacz and Kiran Vodrahalli and Chulayuth Asawaroengchai and Roman Ring and Norbert Kalb and Livio Baldini Soares and Siddhartha Brahma and David Steiner and Tianhe Yu and Fabian Mentzer and Antoine He and Lucas Gonzalez and Bibo Xu and Raphael Lopez Kaufman and Laurent El Shafey and Junhyuk Oh and Tom Hennigan and George van den Driessche and Seth Odoom and Mario Lucic and Becca Roelofs and Sid Lall and Amit Marathe and Betty Chan and Santiago Ontanon and Luheng He and Denis Teplyashin and Jonathan Lai and Phil Crone and Bogdan Damoc and Lewis Ho and Sebastian Riedel and Karel Lenc and Chih-Kuan Yeh and Aakanksha Chowdhery and Yang Xu and Mehran Kazemi and Ehsan Amid and Anastasia Petrushkina and Kevin Swersky and Ali Khodaei and Gowoon Chen and Chris Larkin and Mario Pinto and Geng Yan and Adria Puigdomenech Badia and Piyush Patil and Steven Hansen and Dave Orr and Sebastien M. R. Arnold and Jordan Grimstad and Andrew Dai and Sholto Douglas and Rishika Sinha and Vikas Yadav and Xi Chen and Elena Gribovskaya and Jacob Austin and Jeffrey Zhao and Kaushal Patel and Paul Komarek and Sophia Austin and Sebastian Borgeaud and Linda Friso and Abhimanyu Goyal and Ben Caine and Kris Cao and Da-Woon Chung and Matthew Lamm and Gabe Barth-Maron and Thais Kagohara and Kate Olszewska and Mia Chen and Kaushik Shivakumar and Rishabh Agarwal and Harshal Godhia and Ravi Rajwar and Javier Snaider and Xerxes Dotiwalla and Yuan Liu and Aditya Barua and Victor Ungureanu and Yuan Zhang and Bat-Orgil Batsaikhan and Mateo Wirth and James Qin and Ivo Danihelka and Tulsee Doshi and Martin Chadwick and Jilin Chen and Sanil Jain and Quoc Le and Arjun Kar and Madhu Gurumurthy and Cheng Li and Ruoxin Sang and Fangyu Liu and Lampros Lamprou and Rich Munoz and Nathan Lintz and Harsh Mehta and Heidi Howard and Malcolm Reynolds and Lora Aroyo and Quan Wang and Lorenzo Blanco and Albin Cassirer and Jordan Griffith and Dipanjan Das and Stephan Lee and Jakub Sygnowski and Zach Fisher and James Besley and Richard Powell and Zafarali Ahmed and Dominik Paulus and David Reitter and Zalan Borsos and Rishabh Joshi and Aedan Pope and Steven Hand and Vittorio Selo and Vihan Jain and Nikhil Sethi and Megha Goel and Takaki Makino and Rhys May and Zhen Yang and Johan Schalkwyk and Christina Butterfield and Anja Hauth and Alex Goldin and Will Hawkins and Evan Senter and Sergey Brin and Oliver Woodman and Marvin Ritter and Eric Noland and Minh Giang and Vijay Bolina and Lisa Lee and Tim Blyth and Ian Mackinnon and Machel Reid and Obaid Sarvana and David Silver and Alexander Chen and Lily Wang and Loren Maggiore and Oscar Chang and Nithya Attaluri and Gregory Thornton and Chung-Cheng Chiu and Oskar Bunyan and Nir Levine and Timothy Chung and Evgenii Eltyshev and Xiance Si and Timothy Lillicrap and Demetra Brady and Vaibhav Aggarwal and Boxi Wu and Yuanzhong Xu and Ross McIlroy and Kartikeya Badola and Paramjit Sandhu and Erica Moreira and Wojciech Stokowiec and Ross Hemsley and Dong Li and Alex Tudor and Pranav Shyam and Elahe Rahimtoroghi and Salem Haykal and Pablo Sprechmann and Xiang Zhou and Diana Mincu and Yujia Li and Ravi Addanki and Kalpesh Krishna and Xiao Wu and Alexandre Frechette and Matan Eyal and Allan Dafoe and Dave Lacey and Jay Whang and Thi Avrahami and Ye Zhang and Emanuel Taropa and Hanzhao Lin and Daniel Toyama and Eliza Rutherford and Motoki Sano and HyunJeong Choe and Alex Tomala and Chalence Safranek-Shrader and Nora Kassner and Mantas Pajarskas and Matt Harvey and Sean Sechrist and Meire Fortunato and Christina Lyu and Gamaleldin Elsayed and Chenkai Kuang and James Lottes and Eric Chu and Chao Jia and Chih-Wei Chen and Peter Humphreys and Kate Baumli and Connie Tao and Rajkumar Samuel and Cicero Nogueira dos Santos and Anders Andreassen and Nemanja Rakićević and Dominik Grewe and Aviral Kumar and Stephanie Winkler and Jonathan Caton and Andrew Brock and Sid Dalmia and Hannah Sheahan and Iain Barr and Yingjie Miao and Paul Natsev and Jacob Devlin and Feryal Behbahani and Flavien Prost and Yanhua Sun and Artiom Myaskovsky and Thanumalayan Sankaranarayana Pillai and Dan Hurt and Angeliki Lazaridou and Xi Xiong and Ce Zheng and Fabio Pardo and Xiaowei Li and Dan Horgan and Joe Stanton and Moran Ambar and Fei Xia and Alejandro Lince and Mingqiu Wang and Basil Mustafa and Albert Webson and Hyo Lee and Rohan Anil and Martin Wicke and Timothy Dozat and Abhishek Sinha and Enrique Piqueras and Elahe Dabir and Shyam Upadhyay and Anudhyan Boral and Lisa Anne Hendricks and Corey Fry and Josip Djolonga and Yi Su and Jake Walker and Jane Labanowski and Ronny Huang and Vedant Misra and Jeremy Chen and RJ Skerry-Ryan and Avi Singh and Shruti Rijhwani and Dian Yu and Alex Castro-Ros and Beer Changpinyo and Romina Datta and Sumit Bagri and Arnar Mar Hrafnkelsson and Marcello Maggioni and Daniel Zheng and Yury Sulsky and Shaobo Hou and Tom Le Paine and Antoine Yang and Jason Riesa and Dominika Rogozinska and Dror Marcus and Dalia El Badawy and Qiao Zhang and Luyu Wang and Helen Miller and Jeremy Greer and Lars Lowe Sjos and Azade Nova and Heiga Zen and Rahma Chaabouni and Mihaela Rosca and Jiepu Jiang and Charlie Chen and Ruibo Liu and Tara Sainath and Maxim Krikun and Alex Polozov and Jean-Baptiste Lespiau and Josh Newlan and Zeyncep Cankara and Soo Kwak and Yunhan Xu and Phil Chen and Andy Coenen and Clemens Meyer and Katerina Tsihlas and Ada Ma and Juraj Gottweis and Jinwei Xing and Chenjie Gu and Jin Miao and Christian Frank and Zeynep Cankara and Sanjay Ganapathy and Ishita Dasgupta and Steph Hughes-Fitt and Heng Chen and David Reid and Keran Rong and Hongmin Fan and Joost van Amersfoort and Vincent Zhuang and Aaron Cohen and Shixiang Shane Gu and Anhad Mohananey and Anastasija Ilic and Taylor Tobin and John Wieting and Anna Bortsova and Phoebe Thacker and Emma Wang and Emily Caveness and Justin Chiu and Eren Sezener and Alex Kaskasoli and Steven Baker and Katie Millican and Mohamed Elhawaty and Kostas Aisopos and Carl Lebsack and Nathan Byrd and Hanjun Dai and Wenhao Jia and Matthew Wiethoff and Elnaz Davoodi and Albert Weston and Lakshman Yagati and Arun Ahuja and Isabel Gao and Golan Pundak and Susan Zhang and Michael Azzam and Khe Chai Sim and Sergi Caelles and James Keeling and Abhanshu Sharma and Andy Swing and YaGuang Li and Chenxi Liu and Carrie Grimes Bostock and Yamini Bansal and Zachary Nado and Ankesh Anand and Josh Lipschultz and Abhijit Karmarkar and Lev Proleev and Abe Ittycheriah and Soheil Hassas Yeganeh and George Polovets and Aleksandra Faust and Jiao Sun and Alban Rrustemi and Pen Li and Rakesh Shivanna and Jeremiah Liu and Chris Welty and Federico Lebron and Anirudh Baddepudi and Sebastian Krause and Emilio Parisotto and Radu Soricut and Zheng Xu and Dawn Bloxwich and Melvin Johnson and Behnam Neyshabur and Justin Mao-Jones and Renshen Wang and Vinay Ramasesh and Zaheer Abbas and Arthur Guez and Constant Segal and Duc Dung Nguyen and James Svensson and Le Hou and Sarah York and Kieran Milan and Sophie Bridgers and Wiktor Gworek and Marco Tagliasacchi and James Lee-Thorp and Michael Chang and Alexey Guseynov and Ale Jakse Hartman and Michael Kwong and Ruizhe Zhao and Sheleem Kashem and Elizabeth Cole and Antoine Miech and Richard Tanburn and Mary Phuong and Filip Pavetic and Sebastien Cevey and Ramona Comanescu and Richard Ives and Sherry Yang and Cosmo Du and Bo Li and Zizhao Zhang and Mariko Iinuma and Clara Huiyi Hu and Aurko Roy and Shaan Bijwadia and Zhenkai Zhu and Danilo Martins and Rachel Saputro and Anita Gergely and Steven Zheng and Dawei Jia and Ioannis Antonoglou and Adam Sadovsky and Shane Gu and Yingying Bi and Alek Andreev and Sina Samangooei and Mina Khan and Tomas Kocisky and Angelos Filos and Chintu Kumar and Colton Bishop and Adams Yu and Sarah Hodkinson and Sid Mittal and Premal Shah and Alexandre Moufarek and Yong Cheng and Adam Bloniarz and Jaehoon Lee and Pedram Pejman and Paul Michel and Stephen Spencer and Vladimir Feinberg and Xuehan Xiong and Nikolay Savinov and Charlotte Smith and Siamak Shakeri and Dustin Tran and Mary Chesus and Bernd Bohnet and George Tucker and Tamara von Glehn and Carrie Muir and Yiran Mao and Hideto Kazawa and Ambrose Slone and Kedar Soparkar and Disha Shrivastava and James Cobon-Kerr and Michael Sharman and Jay Pavagadhi and Carlos Araya and Karolis Misiunas and Nimesh Ghelani and Michael Laskin and David Barker and Qiujia Li and Anton Briukhov and Neil Houlsby and Mia Glaese and Balaji Lakshminarayanan and Nathan Schucher and Yunhao Tang and Eli Collins and Hyeontaek Lim and Fangxiaoyu Feng and Adria Recasens and Guangda Lai and Alberto Magni and Nicola De Cao and Aditya Siddhant and Zoe Ashwood and Jordi Orbay and Mostafa Dehghani and Jenny Brennan and Yifan He and Kelvin Xu and Yang Gao and Carl Saroufim and James Molloy and Xinyi Wu and Seb Arnold and Solomon Chang and Julian Schrittwieser and Elena Buchatskaya and Soroush Radpour and Martin Polacek and Skye Giordano and Ankur Bapna and Simon Tokumine and Vincent Hellendoorn and Thibault Sottiaux and Sarah Cogan and Aliaksei Severyn and Mohammad Saleh and Shantanu Thakoor and Laurent Shefey and Siyuan Qiao and Meenu Gaba and Shuo-yiin Chang and Craig Swanson and Biao Zhang and Benjamin Lee and Paul Kishan Rubenstein and Gan Song and Tom Kwiatkowski and Anna Koop and Ajay Kannan and David Kao and Parker Schuh and Axel Stjerngren and Golnaz Ghiasi and Gena Gibson and Luke Vilnis and Ye Yuan and Felipe Tiengo Ferreira and Aishwarya Kamath and Ted Klimenko and Ken Franko and Kefan Xiao and Indro Bhattacharya and Miteyan Patel and Rui Wang and Alex Morris and Robin Strudel and Vivek Sharma and Peter Choy and Sayed Hadi Hashemi and Jessica Landon and Mara Finkelstein and Priya Jhakra and Justin Frye and Megan Barnes and Matthew Mauger and Dennis Daun and Khuslen Baatarsukh and Matthew Tung and Wael Farhan and Henryk Michalewski and Fabio Viola and Felix de Chaumont Quitry and Charline Le Lan and Tom Hudson and Qingze Wang and Felix Fischer and Ivy Zheng and Elspeth White and Anca Dragan and Jean-baptiste Alayrac and Eric Ni and Alexander Pritzel and Adam Iwanicki and Michael Isard and Anna Bulanova and Lukas Zilka and Ethan Dyer and Devendra Sachan and Srivatsan Srinivasan and Hannah Muckenhirn and Honglong Cai and Amol Mandhane and Mukarram Tariq and Jack W. Rae and Gary Wang and Kareem Ayoub and Nicholas FitzGerald and Yao Zhao and Woohyun Han and Chris Alberti and Dan Garrette and Kashyap Krishnakumar and Mai Gimenez and Anselm Levskaya and Daniel Sohn and Josip Matak and Inaki Iturrate and Michael B. Chang and Jackie Xiang and Yuan Cao and Nishant Ranka and Geoff Brown and Adrian Hutter and Vahab Mirrokni and Nanxin Chen and Kaisheng Yao and Zoltan Egyed and Francois Galilee and Tyler Liechty and Praveen Kallakuri and Evan Palmer and Sanjay Ghemawat and Jasmine Liu and David Tao and Chloe Thornton and Tim Green and Mimi Jasarevic and Sharon Lin and Victor Cotruta and Yi-Xuan Tan and Noah Fiedel and Hongkun Yu and Ed Chi and Alexander Neitz and Jens Heitkaemper and Anu Sinha and Denny Zhou and Yi Sun and Charbel Kaed and Brice Hulse and Swaroop Mishra and Maria Georgaki and Sneha Kudugunta and Clement Farabet and Izhak Shafran and Daniel Vlasic and Anton Tsitsulin and Rajagopal Ananthanarayanan and Alen Carin and Guolong Su and Pei Sun and Shashank V and Gabriel Carvajal and Josef Broder and Iulia Comsa and Alena Repina and William Wong and Warren Weilun Chen and Peter Hawkins and Egor Filonov and Lucia Loher and Christoph Hirnschall and Weiyi Wang and Jingchen Ye and Andrea Burns and Hardie Cate and Diana Gage Wright and Federico Piccinini and Lei Zhang and Chu-Cheng Lin and Ionel Gog and Yana Kulizhskaya and Ashwin Sreevatsa and Shuang Song and Luis C. Cobo and Anand Iyer and Chetan Tekur and Guillermo Garrido and Zhuyun Xiao and Rupert Kemp and Huaixiu Steven Zheng and Hui Li and Ananth Agarwal and Christel Ngani and Kati Goshvadi and Rebeca Santamaria-Fernandez and Wojciech Fica and Xinyun Chen and Chris Gorgolewski and Sean Sun and Roopal Garg and Xinyu Ye and S. M. Ali Eslami and Nan Hua and Jon Simon and Pratik Joshi and Yelin Kim and Ian Tenney and Sahitya Potluri and Lam Nguyen Thiet and Quan Yuan and Florian Luisier and Alexandra Chronopoulou and Salvatore Scellato and Praveen Srinivasan and Minmin Chen and Vinod Koverkathu and Valentin Dalibard and Yaming Xu and Brennan Saeta and Keith Anderson and Thibault Sellam and Nick Fernando and Fantine Huot and Junehyuk Jung and Mani Varadarajan and Michael Quinn and Amit Raul and Maigo Le and Ruslan Habalov and Jon Clark and Komal Jalan and Kalesha Bullard and Achintya Singhal and Thang Luong and Boyu Wang and Sujeevan Rajayogam and Julian Eisenschlos and Johnson Jia and Daniel Finchelstein and Alex Yakubovich and Daniel Balle and Michael Fink and Sameer Agarwal and Jing Li and Dj Dvijotham and Shalini Pal and Kai Kang and Jaclyn Konzelmann and Jennifer Beattie and Olivier Dousse and Diane Wu and Remi Crocker and Chen Elkind and Siddhartha Reddy Jonnalagadda and Jong Lee and Dan Holtmann-Rice and Krystal Kallarackal and Rosanne Liu and Denis Vnukov and Neera Vats and Luca Invernizzi and Mohsen Jafari and Huanjie Zhou and Lilly Taylor and Jennifer Prendki and Marcus Wu and Tom Eccles and Tianqi Liu and Kavya Kopparapu and Francoise Beaufays and Christof Angermueller and Andreea Marzoca and Shourya Sarcar and Hilal Dib and Jeff Stanway and Frank Perbet and Nejc Trdin and Rachel Sterneck and Andrey Khorlin and Dinghua Li and Xihui Wu and Sonam Goenka and David Madras and Sasha Goldshtein and Willi Gierke and Tong Zhou and Yaxin Liu and Yannie Liang and Anais White and Yunjie Li and Shreya Singh and Sanaz Bahargam and Mark Epstein and Sujoy Basu and Li Lao and Adnan Ozturel and Carl Crous and Alex Zhai and Han Lu and Zora Tung and Neeraj Gaur and Alanna Walton and Lucas Dixon and Ming Zhang and Amir Globerson and Grant Uy and Andrew Bolt and Olivia Wiles and Milad Nasr and Ilia Shumailov and Marco Selvi and Francesco Piccinno and Ricardo Aguilar and Sara McCarthy and Misha Khalman and Mrinal Shukla and Vlado Galic and John Carpenter and Kevin Villela and Haibin Zhang and Harry Richardson and James Martens and Matko Bosnjak and Shreyas Rammohan Belle and Jeff Seibert and Mahmoud Alnahlawi and Brian McWilliams and Sankalp Singh and Annie Louis and Wen Ding and Dan Popovici and Lenin Simicich and Laura Knight and Pulkit Mehta and Nishesh Gupta and Chongyang Shi and Saaber Fatehi and Jovana Mitrovic and Alex Grills and Joseph Pagadora and Tsendsuren Munkhdalai and Dessie Petrova and Danielle Eisenbud and Zhishuai Zhang and Damion Yates and Bhavishya Mittal and Nilesh Tripuraneni and Yannis Assael and Thomas Brovelli and Prateek Jain and Mihajlo Velimirovic and Canfer Akbulut and Jiaqi Mu and Wolfgang Macherey and Ravin Kumar and Jun Xu and Haroon Qureshi and Gheorghe Comanici and Jeremy Wiesner and Zhitao Gong and Anton Ruddock and Matthias Bauer and Nick Felt and Anirudh GP and Anurag Arnab and Dustin Zelle and Jonas Rothfuss and Bill Rosgen and Ashish Shenoy and Bryan Seybold and Xinjian Li and Jayaram Mudigonda and Goker Erdogan and Jiawei Xia and Jiri Simsa and Andrea Michi and Yi Yao and Christopher Yew and Steven Kan and Isaac Caswell and Carey Radebaugh and Andre Elisseeff and Pedro Valenzuela and Kay McKinney and Kim Paterson and Albert Cui and Eri Latorre-Chimoto and Solomon Kim and William Zeng and Ken Durden and Priya Ponnapalli and Tiberiu Sosea and Christopher A. Choquette-Choo and James Manyika and Brona Robenek and Harsha Vashisht and Sebastien Pereira and Hoi Lam and Marko Velic and Denese Owusu-Afriyie and Katherine Lee and Tolga Bolukbasi and Alicia Parrish and Shawn Lu and Jane Park and Balaji Venkatraman and Alice Talbert and Lambert Rosique and Yuchung Cheng and Andrei Sozanschi and Adam Paszke and Praveen Kumar and Jessica Austin and Lu Li and Khalid Salama and Bartek Perz and Wooyeol Kim and Nandita Dukkipati and Anthony Baryshnikov and Christos Kaplanis and XiangHai Sheng and Yuri Chervonyi and Caglar Unlu and Diego de Las Casas and Harry Askham and Kathryn Tunyasuvunakool and Felix Gimeno and Siim Poder and Chester Kwak and Matt Miecnikowski and Vahab Mirrokni and Alek Dimitriev and Aaron Parisi and Dangyi Liu and Tomy Tsai and Toby Shevlane and Christina Kouridi and Drew Garmon and Adrian Goedeckemeyer and Adam R. Brown and Anitha Vijayakumar and Ali Elqursh and Sadegh Jazayeri and Jin Huang and Sara Mc Carthy and Jay Hoover and Lucy Kim and Sandeep Kumar and Wei Chen and Courtney Biles and Garrett Bingham and Evan Rosen and Lisa Wang and Qijun Tan and David Engel and Francesco Pongetti and Dario de Cesare and Dongseong Hwang and Lily Yu and Jennifer Pullman and Srini Narayanan and Kyle Levin and Siddharth Gopal and Megan Li and Asaf Aharoni and Trieu Trinh and Jessica Lo and Norman Casagrande and Roopali Vij and Loic Matthey and Bramandia Ramadhana and Austin Matthews and CJ Carey and Matthew Johnson and Kremena Goranova and Rohin Shah and Shereen Ashraf and Kingshuk Dasgupta and Rasmus Larsen and Yicheng Wang and Manish Reddy Vuyyuru and Chong Jiang and Joana Ijazi and Kazuki Osawa and Celine Smith and Ramya Sree Boppana and Taylan Bilal and Yuma Koizumi and Ying Xu and Yasemin Altun and Nir Shabat and Ben Bariach and Alex Korchemniy and Kiam Choo and Olaf Ronneberger and Chimezie Iwuanyanwu and Shubin Zhao and David Soergel and Cho-Jui Hsieh and Irene Cai and Shariq Iqbal and Martin Sundermeyer and Zhe Chen and Elie Bursztein and Chaitanya Malaviya and Fadi Biadsy and Prakash Shroff and Inderjit Dhillon and Tejasi Latkar and Chris Dyer and Hannah Forbes and Massimo Nicosia and Vitaly Nikolaev and Somer Greene and Marin Georgiev and Pidong Wang and Nina Martin and Hanie Sedghi and John Zhang and Praseem Banzal and Doug Fritz and Vikram Rao and Xuezhi Wang and Jiageng Zhang and Viorica Patraucean and Dayou Du and Igor Mordatch and Ivan Jurin and Lewis Liu and Ayush Dubey and Abhi Mohan and Janek Nowakowski and Vlad-Doru Ion and Nan Wei and Reiko Tojo and Maria Abi Raad and Drew A. Hudson and Vaishakh Keshava and Shubham Agrawal and Kevin Ramirez and Zhichun Wu and Hoang Nguyen and Ji Liu and Madhavi Sewak and Bryce Petrini and DongHyun Choi and Ivan Philips and Ziyue Wang and Ioana Bica and Ankush Garg and Jarek Wilkiewicz and Priyanka Agrawal and Xiaowei Li and Danhao Guo and Emily Xue and Naseer Shaik and Andrew Leach and Sadh MNM Khan and Julia Wiesinger and Sammy Jerome and Abhishek Chakladar and Alek Wenjiao Wang and Tina Ornduff and Folake Abu and Alireza Ghaffarkhah and Marcus Wainwright and Mario Cortes and Frederick Liu and Joshua Maynez and Andreas Terzis and Pouya Samangouei and Riham Mansour and Tomasz Kępa and François-Xavier Aubet and Anton Algymr and Dan Banica and Agoston Weisz and Andras Orban and Alexandre Senges and Ewa Andrejczuk and Mark Geller and Niccolo Dal Santo and Valentin Anklin and Majd Al Merey and Martin Baeuml and Trevor Strohman and Junwen Bai and Slav Petrov and Yonghui Wu and Demis Hassabis and Koray Kavukcuoglu and Jeff Dean and Oriol Vinyals},
      year={2024},
      eprint={2403.05530},
      archivePrefix={arXiv},
      primaryClass={cs.CL},
      url={https://arxiv.org/abs/2403.05530}, 
}

@misc{gpt42024,
title={{GPT-4 Technical Report}}, 
      author={OpenAI and Josh Achiam and Steven Adler and Sandhini Agarwal and Lama Ahmad and Ilge Akkaya and Florencia Leoni Aleman and Diogo Almeida and Janko Altenschmidt and Sam Altman and Shyamal Anadkat and Red Avila and Igor Babuschkin and Suchir Balaji and Valerie Balcom and Paul Baltescu and Haiming Bao and Mohammad Bavarian and Jeff Belgum and Irwan Bello and Jake Berdine and Gabriel Bernadett-Shapiro and Christopher Berner and Lenny Bogdonoff and Oleg Boiko and Madelaine Boyd and Anna-Luisa Brakman and Greg Brockman and Tim Brooks and Miles Brundage and Kevin Button and Trevor Cai and Rosie Campbell and Andrew Cann and Brittany Carey and Chelsea Carlson and Rory Carmichael and Brooke Chan and Che Chang and Fotis Chantzis and Derek Chen and Sully Chen and Ruby Chen and Jason Chen and Mark Chen and Ben Chess and Chester Cho and Casey Chu and Hyung Won Chung and Dave Cummings and Jeremiah Currier and Yunxing Dai and Cory Decareaux and Thomas Degry and Noah Deutsch and Damien Deville and Arka Dhar and David Dohan and Steve Dowling and Sheila Dunning and Adrien Ecoffet and Atty Eleti and Tyna Eloundou and David Farhi and Liam Fedus and Niko Felix and Simón Posada Fishman and Juston Forte and Isabella Fulford and Leo Gao and Elie Georges and Christian Gibson and Vik Goel and Tarun Gogineni and Gabriel Goh and Rapha Gontijo-Lopes and Jonathan Gordon and Morgan Grafstein and Scott Gray and Ryan Greene and Joshua Gross and Shixiang Shane Gu and Yufei Guo and Chris Hallacy and Jesse Han and Jeff Harris and Yuchen He and Mike Heaton and Johannes Heidecke and Chris Hesse and Alan Hickey and Wade Hickey and Peter Hoeschele and Brandon Houghton and Kenny Hsu and Shengli Hu and Xin Hu and Joost Huizinga and Shantanu Jain and Shawn Jain and Joanne Jang and Angela Jiang and Roger Jiang and Haozhun Jin and Denny Jin and Shino Jomoto and Billie Jonn and Heewoo Jun and Tomer Kaftan and Łukasz Kaiser and Ali Kamali and Ingmar Kanitscheider and Nitish Shirish Keskar and Tabarak Khan and Logan Kilpatrick and Jong Wook Kim and Christina Kim and Yongjik Kim and Jan Hendrik Kirchner and Jamie Kiros and Matt Knight and Daniel Kokotajlo and Łukasz Kondraciuk and Andrew Kondrich and Aris Konstantinidis and Kyle Kosic and Gretchen Krueger and Vishal Kuo and Michael Lampe and Ikai Lan and Teddy Lee and Jan Leike and Jade Leung and Daniel Levy and Chak Ming Li and Rachel Lim and Molly Lin and Stephanie Lin and Mateusz Litwin and Theresa Lopez and Ryan Lowe and Patricia Lue and Anna Makanju and Kim Malfacini and Sam Manning and Todor Markov and Yaniv Markovski and Bianca Martin and Katie Mayer and Andrew Mayne and Bob McGrew and Scott Mayer McKinney and Christine McLeavey and Paul McMillan and Jake McNeil and David Medina and Aalok Mehta and Jacob Menick and Luke Metz and Andrey Mishchenko and Pamela Mishkin and Vinnie Monaco and Evan Morikawa and Daniel Mossing and Tong Mu and Mira Murati and Oleg Murk and David Mély and Ashvin Nair and Reiichiro Nakano and Rajeev Nayak and Arvind Neelakantan and Richard Ngo and Hyeonwoo Noh and Long Ouyang and Cullen O'Keefe and Jakub Pachocki and Alex Paino and Joe Palermo and Ashley Pantuliano and Giambattista Parascandolo and Joel Parish and Emy Parparita and Alex Passos and Mikhail Pavlov and Andrew Peng and Adam Perelman and Filipe de Avila Belbute Peres and Michael Petrov and Henrique Ponde de Oliveira Pinto and Michael and Pokorny and Michelle Pokrass and Vitchyr H. Pong and Tolly Powell and Alethea Power and Boris Power and Elizabeth Proehl and Raul Puri and Alec Radford and Jack Rae and Aditya Ramesh and Cameron Raymond and Francis Real and Kendra Rimbach and Carl Ross and Bob Rotsted and Henri Roussez and Nick Ryder and Mario Saltarelli and Ted Sanders and Shibani Santurkar and Girish Sastry and Heather Schmidt and David Schnurr and John Schulman and Daniel Selsam and Kyla Sheppard and Toki Sherbakov and Jessica Shieh and Sarah Shoker and Pranav Shyam and Szymon Sidor and Eric Sigler and Maddie Simens and Jordan Sitkin and Katarina Slama and Ian Sohl and Benjamin Sokolowsky and Yang Song and Natalie Staudacher and Felipe Petroski Such and Natalie Summers and Ilya Sutskever and Jie Tang and Nikolas Tezak and Madeleine B. Thompson and Phil Tillet and Amin Tootoonchian and Elizabeth Tseng and Preston Tuggle and Nick Turley and Jerry Tworek and Juan Felipe Cerón Uribe and Andrea Vallone and Arun Vijayvergiya and Chelsea Voss and Carroll Wainwright and Justin Jay Wang and Alvin Wang and Ben Wang and Jonathan Ward and Jason Wei and CJ Weinmann and Akila Welihinda and Peter Welinder and Jiayi Weng and Lilian Weng and Matt Wiethoff and Dave Willner and Clemens Winter and Samuel Wolrich and Hannah Wong and Lauren Workman and Sherwin Wu and Jeff Wu and Michael Wu and Kai Xiao and Tao Xu and Sarah Yoo and Kevin Yu and Qiming Yuan and Wojciech Zaremba and Rowan Zellers and Chong Zhang and Marvin Zhang and Shengjia Zhao and Tianhao Zheng and Juntang Zhuang and William Zhuk and Barret Zoph},
      year={2024},
      eprint={2303.08774},
      archivePrefix={arXiv},
      primaryClass={cs.CL},
      url={https://arxiv.org/abs/2303.08774}, 
}

@misc{llama2024,
      title={{The Llama 3 Herd of Models}}, 
      author={Aaron Grattafiori and Abhimanyu Dubey and Abhinav Jauhri and Abhinav Pandey and Abhishek Kadian and Ahmad Al-Dahle and Aiesha Letman and Akhil Mathur and Alan Schelten and Alex Vaughan and Amy Yang and Angela Fan and Anirudh Goyal and Anthony Hartshorn and Aobo Yang and Archi Mitra and Archie Sravankumar and Artem Korenev and Arthur Hinsvark and Arun Rao and Aston Zhang and Aurelien Rodriguez and Austen Gregerson and Ava Spataru and Baptiste Roziere and Bethany Biron and Binh Tang and Bobbie Chern and Charlotte Caucheteux and Chaya Nayak and Chloe Bi and Chris Marra and Chris McConnell and Christian Keller and Christophe Touret and Chunyang Wu and Corinne Wong and Cristian Canton Ferrer and Cyrus Nikolaidis and Damien Allonsius and Daniel Song and Danielle Pintz and Danny Livshits and Danny Wyatt and David Esiobu and Dhruv Choudhary and Dhruv Mahajan and Diego Garcia-Olano and Diego Perino and Dieuwke Hupkes and Egor Lakomkin and Ehab AlBadawy and Elina Lobanova and Emily Dinan and Eric Michael Smith and Filip Radenovic and Francisco Guzmán and Frank Zhang and Gabriel Synnaeve and Gabrielle Lee and Georgia Lewis Anderson and Govind Thattai and Graeme Nail and Gregoire Mialon and Guan Pang and Guillem Cucurell and Hailey Nguyen and Hannah Korevaar and Hu Xu and Hugo Touvron and Iliyan Zarov and Imanol Arrieta Ibarra and Isabel Kloumann and Ishan Misra and Ivan Evtimov and Jack Zhang and Jade Copet and Jaewon Lee and Jan Geffert and Jana Vranes and Jason Park and Jay Mahadeokar and Jeet Shah and Jelmer van der Linde and Jennifer Billock and Jenny Hong and Jenya Lee and Jeremy Fu and Jianfeng Chi and Jianyu Huang and Jiawen Liu and Jie Wang and Jiecao Yu and Joanna Bitton and Joe Spisak and Jongsoo Park and Joseph Rocca and Joshua Johnstun and Joshua Saxe and Junteng Jia and Kalyan Vasuden Alwala and Karthik Prasad and Kartikeya Upasani and Kate Plawiak and Ke Li and Kenneth Heafield and Kevin Stone and Khalid El-Arini and Krithika Iyer and Kshitiz Malik and Kuenley Chiu and Kunal Bhalla and Kushal Lakhotia and Lauren Rantala-Yeary and Laurens van der Maaten and Lawrence Chen and Liang Tan and Liz Jenkins and Louis Martin and Lovish Madaan and Lubo Malo and Lukas Blecher and Lukas Landzaat and Luke de Oliveira and Madeline Muzzi and Mahesh Pasupuleti and Mannat Singh and Manohar Paluri and Marcin Kardas and Maria Tsimpoukelli and Mathew Oldham and Mathieu Rita and Maya Pavlova and Melanie Kambadur and Mike Lewis and Min Si and Mitesh Kumar Singh and Mona Hassan and Naman Goyal and Narjes Torabi and Nikolay Bashlykov and Nikolay Bogoychev and Niladri Chatterji and Ning Zhang and Olivier Duchenne and Onur Çelebi and Patrick Alrassy and Pengchuan Zhang and Pengwei Li and Petar Vasic and Peter Weng and Prajjwal Bhargava and Pratik Dubal and Praveen Krishnan and Punit Singh Koura and Puxin Xu and Qing He and Qingxiao Dong and Ragavan Srinivasan and Raj Ganapathy and Ramon Calderer and Ricardo Silveira Cabral and Robert Stojnic and Roberta Raileanu and Rohan Maheswari and Rohit Girdhar and Rohit Patel and Romain Sauvestre and Ronnie Polidoro and Roshan Sumbaly and Ross Taylor and Ruan Silva and Rui Hou and Rui Wang and Saghar Hosseini and Sahana Chennabasappa and Sanjay Singh and Sean Bell and Seohyun Sonia Kim and Sergey Edunov and Shaoliang Nie and Sharan Narang and Sharath Raparthy and Sheng Shen and Shengye Wan and Shruti Bhosale and Shun Zhang and Simon Vandenhende and Soumya Batra and Spencer Whitman and Sten Sootla and Stephane Collot and Suchin Gururangan and Sydney Borodinsky and Tamar Herman and Tara Fowler and Tarek Sheasha and Thomas Georgiou and Thomas Scialom and Tobias Speckbacher and Todor Mihaylov and Tong Xiao and Ujjwal Karn and Vedanuj Goswami and Vibhor Gupta and Vignesh Ramanathan and Viktor Kerkez and Vincent Gonguet and Virginie Do and Vish Vogeti and Vítor Albiero and Vladan Petrovic and Weiwei Chu and Wenhan Xiong and Wenyin Fu and Whitney Meers and Xavier Martinet and Xiaodong Wang and Xiaofang Wang and Xiaoqing Ellen Tan and Xide Xia and Xinfeng Xie and Xuchao Jia and Xuewei Wang and Yaelle Goldschlag and Yashesh Gaur and Yasmine Babaei and Yi Wen and Yiwen Song and Yuchen Zhang and Yue Li and Yuning Mao and Zacharie Delpierre Coudert and Zheng Yan and Zhengxing Chen and Zoe Papakipos and Aaditya Singh and Aayushi Srivastava and Abha Jain and Adam Kelsey and Adam Shajnfeld and Adithya Gangidi and Adolfo Victoria and Ahuva Goldstand and Ajay Menon and Ajay Sharma and Alex Boesenberg and Alexei Baevski and Allie Feinstein and Amanda Kallet and Amit Sangani and Amos Teo and Anam Yunus and Andrei Lupu and Andres Alvarado and Andrew Caples and Andrew Gu and Andrew Ho and Andrew Poulton and Andrew Ryan and Ankit Ramchandani and Annie Dong and Annie Franco and Anuj Goyal and Aparajita Saraf and Arkabandhu Chowdhury and Ashley Gabriel and Ashwin Bharambe and Assaf Eisenman and Azadeh Yazdan and Beau James and Ben Maurer and Benjamin Leonhardi and Bernie Huang and Beth Loyd and Beto De Paola and Bhargavi Paranjape and Bing Liu and Bo Wu and Boyu Ni and Braden Hancock and Bram Wasti and Brandon Spence and Brani Stojkovic and Brian Gamido and Britt Montalvo and Carl Parker and Carly Burton and Catalina Mejia and Ce Liu and Changhan Wang and Changkyu Kim and Chao Zhou and Chester Hu and Ching-Hsiang Chu and Chris Cai and Chris Tindal and Christoph Feichtenhofer and Cynthia Gao and Damon Civin and Dana Beaty and Daniel Kreymer and Daniel Li and David Adkins and David Xu and Davide Testuggine and Delia David and Devi Parikh and Diana Liskovich and Didem Foss and Dingkang Wang and Duc Le and Dustin Holland and Edward Dowling and Eissa Jamil and Elaine Montgomery and Eleonora Presani and Emily Hahn and Emily Wood and Eric-Tuan Le and Erik Brinkman and Esteban Arcaute and Evan Dunbar and Evan Smothers and Fei Sun and Felix Kreuk and Feng Tian and Filippos Kokkinos and Firat Ozgenel and Francesco Caggioni and Frank Kanayet and Frank Seide and Gabriela Medina Florez and Gabriella Schwarz and Gada Badeer and Georgia Swee and Gil Halpern and Grant Herman and Grigory Sizov and Guangyi and Zhang and Guna Lakshminarayanan and Hakan Inan and Hamid Shojanazeri and Han Zou and Hannah Wang and Hanwen Zha and Haroun Habeeb and Harrison Rudolph and Helen Suk and Henry Aspegren and Hunter Goldman and Hongyuan Zhan and Ibrahim Damlaj and Igor Molybog and Igor Tufanov and Ilias Leontiadis and Irina-Elena Veliche and Itai Gat and Jake Weissman and James Geboski and James Kohli and Janice Lam and Japhet Asher and Jean-Baptiste Gaya and Jeff Marcus and Jeff Tang and Jennifer Chan and Jenny Zhen and Jeremy Reizenstein and Jeremy Teboul and Jessica Zhong and Jian Jin and Jingyi Yang and Joe Cummings and Jon Carvill and Jon Shepard and Jonathan McPhie and Jonathan Torres and Josh Ginsburg and Junjie Wang and Kai Wu and Kam Hou U and Karan Saxena and Kartikay Khandelwal and Katayoun Zand and Kathy Matosich and Kaushik Veeraraghavan and Kelly Michelena and Keqian Li and Kiran Jagadeesh and Kun Huang and Kunal Chawla and Kyle Huang and Lailin Chen and Lakshya Garg and Lavender A and Leandro Silva and Lee Bell and Lei Zhang and Liangpeng Guo and Licheng Yu and Liron Moshkovich and Luca Wehrstedt and Madian Khabsa and Manav Avalani and Manish Bhatt and Martynas Mankus and Matan Hasson and Matthew Lennie and Matthias Reso and Maxim Groshev and Maxim Naumov and Maya Lathi and Meghan Keneally and Miao Liu and Michael L. Seltzer and Michal Valko and Michelle Restrepo and Mihir Patel and Mik Vyatskov and Mikayel Samvelyan and Mike Clark and Mike Macey and Mike Wang and Miquel Jubert Hermoso and Mo Metanat and Mohammad Rastegari and Munish Bansal and Nandhini Santhanam and Natascha Parks and Natasha White and Navyata Bawa and Nayan Singhal and Nick Egebo and Nicolas Usunier and Nikhil Mehta and Nikolay Pavlovich Laptev and Ning Dong and Norman Cheng and Oleg Chernoguz and Olivia Hart and Omkar Salpekar and Ozlem Kalinli and Parkin Kent and Parth Parekh and Paul Saab and Pavan Balaji and Pedro Rittner and Philip Bontrager and Pierre Roux and Piotr Dollar and Polina Zvyagina and Prashant Ratanchandani and Pritish Yuvraj and Qian Liang and Rachad Alao and Rachel Rodriguez and Rafi Ayub and Raghotham Murthy and Raghu Nayani and Rahul Mitra and Rangaprabhu Parthasarathy and Raymond Li and Rebekkah Hogan and Robin Battey and Rocky Wang and Russ Howes and Ruty Rinott and Sachin Mehta and Sachin Siby and Sai Jayesh Bondu and Samyak Datta and Sara Chugh and Sara Hunt and Sargun Dhillon and Sasha Sidorov and Satadru Pan and Saurabh Mahajan and Saurabh Verma and Seiji Yamamoto and Sharadh Ramaswamy and Shaun Lindsay and Shaun Lindsay and Sheng Feng and Shenghao Lin and Shengxin Cindy Zha and Shishir Patil and Shiva Shankar and Shuqiang Zhang and Shuqiang Zhang and Sinong Wang and Sneha Agarwal and Soji Sajuyigbe and Soumith Chintala and Stephanie Max and Stephen Chen and Steve Kehoe and Steve Satterfield and Sudarshan Govindaprasad and Sumit Gupta and Summer Deng and Sungmin Cho and Sunny Virk and Suraj Subramanian and Sy Choudhury and Sydney Goldman and Tal Remez and Tamar Glaser and Tamara Best and Thilo Koehler and Thomas Robinson and Tianhe Li and Tianjun Zhang and Tim Matthews and Timothy Chou and Tzook Shaked and Varun Vontimitta and Victoria Ajayi and Victoria Montanez and Vijai Mohan and Vinay Satish Kumar and Vishal Mangla and Vlad Ionescu and Vlad Poenaru and Vlad Tiberiu Mihailescu and Vladimir Ivanov and Wei Li and Wenchen Wang and Wenwen Jiang and Wes Bouaziz and Will Constable and Xiaocheng Tang and Xiaojian Wu and Xiaolan Wang and Xilun Wu and Xinbo Gao and Yaniv Kleinman and Yanjun Chen and Ye Hu and Ye Jia and Ye Qi and Yenda Li and Yilin Zhang and Ying Zhang and Yossi Adi and Youngjin Nam and Yu and Wang and Yu Zhao and Yuchen Hao and Yundi Qian and Yunlu Li and Yuzi He and Zach Rait and Zachary DeVito and Zef Rosnbrick and Zhaoduo Wen and Zhenyu Yang and Zhiwei Zhao and Zhiyu Ma},
      year={2024},
      eprint={2407.21783},
      archivePrefix={arXiv},
      primaryClass={cs.AI},
      url={https://arxiv.org/abs/2407.21783}, 
}

@misc{qwen251m2025,
      title={{Qwen2.5-1M Technical Report}}, 
      author={An Yang and Bowen Yu and Chengyuan Li and Dayiheng Liu and Fei Huang and Haoyan Huang and Jiandong Jiang and Jianhong Tu and Jianwei Zhang and Jingren Zhou and Junyang Lin and Kai Dang and Kexin Yang and Le Yu and Mei Li and Minmin Sun and Qin Zhu and Rui Men and Tao He and Weijia Xu and Wenbiao Yin and Wenyuan Yu and Xiafei Qiu and Xingzhang Ren and Xinlong Yang and Yong Li and Zhiying Xu and Zipeng Zhang},
      year={2025},
      eprint={2501.15383},
      archivePrefix={arXiv},
      primaryClass={cs.CL},
      url={https://arxiv.org/abs/2501.15383}, 
}

@misc{qwen252025,
      title={{Qwen2.5 Technical Report}}, 
      author={Qwen and : and An Yang and Baosong Yang and Beichen Zhang and Binyuan Hui and Bo Zheng and Bowen Yu and Chengyuan Li and Dayiheng Liu and Fei Huang and Haoran Wei and Huan Lin and Jian Yang and Jianhong Tu and Jianwei Zhang and Jianxin Yang and Jiaxi Yang and Jingren Zhou and Junyang Lin and Kai Dang and Keming Lu and Keqin Bao and Kexin Yang and Le Yu and Mei Li and Mingfeng Xue and Pei Zhang and Qin Zhu and Rui Men and Runji Lin and Tianhao Li and Tianyi Tang and Tingyu Xia and Xingzhang Ren and Xuancheng Ren and Yang Fan and Yang Su and Yichang Zhang and Yu Wan and Yuqiong Liu and Zeyu Cui and Zhenru Zhang and Zihan Qiu},
      year={2025},
      eprint={2412.15115},
      archivePrefix={arXiv},
      primaryClass={cs.CL},
      url={https://arxiv.org/abs/2412.15115}, 
}

@inproceedings{retrievalaugmented2020,
 author = {Lewis, Patrick and Perez, Ethan and Piktus, Aleksandra and Petroni, Fabio and Karpukhin, Vladimir and Goyal, Naman and K\"{u}ttler, Heinrich and Lewis, Mike and Yih, Wen-tau and Rockt\"{a}schel, Tim and Riedel, Sebastian and Kiela, Douwe},
 booktitle = {Advances in Neural Information Processing Systems},
 editor = {H. Larochelle and M. Ranzato and R. Hadsell and M.F. Balcan and H. Lin},
 pages = {9459--9474},
 publisher = {Curran Associates, Inc.},
 title = {{Retrieval-Augmented Generation for Knowledge-Intensive NLP Tasks}},
 url = {https://proceedings.neurips.cc/paper_files/paper/2020/file/6b493230205f780e1bc26945df7481e5-Paper.pdf},
 volume = {33},
 year = {2020},
 address = {Virtual},
}

@inproceedings{generative2023,
author = {Park, Joon Sung and O'Brien, Joseph and Cai, Carrie Jun and Morris, Meredith Ringel and Liang, Percy and Bernstein, Michael S.},
title = {{Generative Agents: Interactive Simulacra of Human Behavior}},
year = {2023},
isbn = {9798400701320},
publisher = {Association for Computing Machinery},
address = {New York, NY, USA},
url = {https://doi.org/10.1145/3586183.3606763},
doi = {10.1145/3586183.3606763},
booktitle = {Proceedings of the 36th Annual ACM Symposium on User Interface Software and Technology},
articleno = {2},
numpages = {22},
location = {San Francisco, CA, USA},
series = {UIST '23}
}

@article{survey2024a,
title={{A survey on large language model based autonomous agents}},
   volume={18},
   ISSN={2095-2236},
   url={http://dx.doi.org/10.1007/s11704-024-40231-1},
   DOI={10.1007/s11704-024-40231-1},
   journal={Frontiers of Computer Science},
   publisher={Springer Science and Business Media LLC},
   author={Wang, Lei and Ma, Chen and Feng, Xueyang and Zhang, Zeyu and Yang, Hao and Zhang, Jingsen and Chen, Zhiyuan and Tang, Jiakai and Chen, Xu and Lin, Yankai and Zhao, Wayne Xin and Wei, Zhewei and Wen, Jirong},
   year={2024},
   pages = {186345},
   number = {6},
   month=Mar
}

@inproceedings{ruler2024,
title={{RULER: What{\textquoteright}s the Real Context Size of Your Long-Context Language Models?}},
author={Cheng-Ping Hsieh and Simeng Sun and Samuel Kriman and Shantanu Acharya and Dima Rekesh and Fei Jia and Boris Ginsburg},
booktitle={First Conference on Language Modeling},
year={2024},
url={https://openreview.net/forum?id=kIoBbc76Sy},
publisher={OpenReview.net},
address={Philadelphia, Pennsylvania, United States},
}

@misc{lm-eval-harness2024,
author       = {Gao, Leo and Tow, Jonathan and Abbasi, Baber and Biderman, Stella and Black, Sid and DiPofi, Anthony and Foster, Charles and Golding, Laurence and Hsu, Jeffrey and Le Noac'h, Alain and Li, Haonan and McDonell, Kyle and Muennighoff, Niklas and Ociepa, Chris and Phang, Jason and Reynolds, Laria and Schoelkopf, Hailey and Skowron, Aviya and Sutawika, Lintang and Tang, Eric and Thite, Anish and Wang, Ben and Wang, Kevin and Zou, Andy},
  title        = {{The Language Model Evaluation Harness}},
  month        = 07,
  year         = 2024,
  publisher    = {Zenodo},
  version      = {v0.4.3},
  doi          = {10.5281/zenodo.12608602},
  url          = {https://zenodo.org/records/12608602}
}

@misc{yang2025qwen3technicalreport,
title={{Qwen3 Technical Report}}, 
      author={An Yang and Anfeng Li and Baosong Yang and Beichen Zhang and Binyuan Hui and Bo Zheng and Bowen Yu and Chang Gao and Chengen Huang and Chenxu Lv and Chujie Zheng and Dayiheng Liu and Fan Zhou and Fei Huang and Feng Hu and Hao Ge and Haoran Wei and Huan Lin and Jialong Tang and Jian Yang and Jianhong Tu and Jianwei Zhang and Jianxin Yang and Jiaxi Yang and Jing Zhou and Jingren Zhou and Junyang Lin and Kai Dang and Keqin Bao and Kexin Yang and Le Yu and Lianghao Deng and Mei Li and Mingfeng Xue and Mingze Li and Pei Zhang and Peng Wang and Qin Zhu and Rui Men and Ruize Gao and Shixuan Liu and Shuang Luo and Tianhao Li and Tianyi Tang and Wenbiao Yin and Xingzhang Ren and Xinyu Wang and Xinyu Zhang and Xuancheng Ren and Yang Fan and Yang Su and Yichang Zhang and Yinger Zhang and Yu Wan and Yuqiong Liu and Zekun Wang and Zeyu Cui and Zhenru Zhang and Zhipeng Zhou and Zihan Qiu},
      year={2025},
      eprint={2505.09388},
      archivePrefix={arXiv},
      primaryClass={cs.CL},
      url={https://arxiv.org/abs/2505.09388}, 
}

@inproceedings{peng2026yarnefficientcontextwindow,
 author = {Peng, Bowen and Quesnelle, Jeffrey and Fan, Honglu and Shippole, Enrico},
 booktitle = {International Conference on Learning Representations},
 editor = {B. Kim and Y. Yue and S. Chaudhuri and K. Fragkiadaki and M. Khan and Y. Sun},
 pages = {31932--31951},
 title = {{YaRN: Efficient Context Window Extension of Large Language Models}},
 url = {https://proceedings.iclr.cc/paper_files/paper/2024/file/874a4d89f2d04b4bcf9a2c19545cf040-Paper-Conference.pdf},
 volume = {2024},
 address = {Vienna, Austria},
 year = {2024}
}

@misc{peng2026grkv,
  title={{GRKV: Global Regression for Training-Free KV Cache Compression in Long-Context LLMs}}, 
  author={Junjie Peng and You Wu and Haoyi Wu and Jialong Han and Xiaohua Xie and Kewei Tu and Jianhuang Lai},
  year={2026},
  eprint={2605.31105},
  archivePrefix={arXiv},
  primaryClass={cs.CL},
  url={https://arxiv.org/abs/2605.31105}, 
}

@inproceedings{zhu-etal-2026-ojakv,
    title = "{{OjaKV: Context-Aware Online Low-Rank KV Cache Compression}}",
    author = "Zhu, Yuxuan  and
      Yang, David H.  and
      Amiri, Mohammad Mohammadi  and
      Murugesan, Keerthiram  and
      Pedapati, Tejaswini  and
      Chen, Pin-Yu",
    editor = "Liakata, Maria  and
      Moreira, Viviane P.  and
      Zhang, Jiajun  and
      Jurgens, David",
    booktitle = "Findings of the {A}ssociation for {C}omputational {L}inguistics: {ACL} 2026",
    month = jul,
    year = "2026",
    address = "San Diego, California, United States",
    publisher = "Association for Computational Linguistics",
    url = "https://aclanthology.org/2026.findings-acl.494/",
    doi = "10.18653/v1/2026.findings-acl.494",
    pages = "10161--10178",
    ISBN = "979-8-89176-395-1"
}

@inproceedings{yang-etal-2024-pyramidinfer,
    title = "{{PyramidInfer: Pyramid KV Cache Compression for High-throughput LLM Inference}}",
    author = "Yang, Dongjie  and
      Han, Xiaodong  and
      Gao, Yan  and
      Hu, Yao  and
      Zhang, Shilin  and
      Zhao, Hai",
    editor = "Ku, Lun-Wei  and
      Martins, Andre  and
      Srikumar, Vivek",
    booktitle = "Findings of the Association for Computational Linguistics: ACL 2024",
    month = aug,
    year = "2024",
    address = "Bangkok, Thailand",
    publisher = "Association for Computational Linguistics",
    url = "https://aclanthology.org/2024.findings-acl.195/",
    doi = "10.18653/v1/2024.findings-acl.195",
    pages = "3258--3270"
}

@inproceedings{joshi-etal-2025-tada,
    title = "{{TaDA: Training-free recipe for Decoding with Adaptive KV Cache Compression and Mean-centering}}",
    author = "Joshi, Vinay  and
      Brahma, Pratik Prabhanjan  and
      Liu, Zicheng  and
      Barsoum, Emad",
    editor = "Rehm, Georg  and
      Li, Yunyao",
    booktitle = "Proceedings of the 63rd Annual Meeting of the Association for Computational Linguistics (Volume 6: Industry Track)",
    month = jul,
    year = "2025",
    address = "Vienna, Austria",
    publisher = "Association for Computational Linguistics",
    url = "https://aclanthology.org/2025.acl-industry.101/",
    doi = "10.18653/v1/2025.acl-industry.101",
    pages = "1435--1443",
    ISBN = "979-8-89176-288-6"
}

@article{lee-etal-2025-tale,
    title = "{{TALE: Token-Adaptive Low-Rank KVCache Approximation with Reconstruction Elimination}}",
    author = "Lee, Jaeseong  and
      Hwang, Seung-won  and
      Qiao, Aurick  and
      Campos, Daniel  and
      Yao, Zhewei  and
      He, Yuxiong",
    journal = "Transactions of the Association for Computational Linguistics",
    volume = "13",
    year = "2025",
    address = "Cambridge, MA",
    publisher = "MIT Press",
    url = "https://aclanthology.org/2025.tacl-1.59/",
    doi = "10.1162/tacl.a.39",
    pages = "1298--1318"
}

@inproceedings{saxena-etal-2024-eigen,
    title = "{{Eigen Attention: Attention in Low-Rank Space for KV Cache Compression}}",
    author = "Saxena, Utkarsh  and
      Saha, Gobinda  and
      Choudhary, Sakshi  and
      Roy, Kaushik",
    editor = "Al-Onaizan, Yaser  and
      Bansal, Mohit  and
      Chen, Yun-Nung",
    booktitle = "Findings of the Association for Computational Linguistics: EMNLP 2024",
    month = nov,
    year = "2024",
    address = "Miami, Florida, USA",
    publisher = "Association for Computational Linguistics",
    url = "https://aclanthology.org/2024.findings-emnlp.899/",
    doi = "10.18653/v1/2024.findings-emnlp.899",
    pages = "15332--15344"
}

@inproceedings{bai-etal-2025-longbench,
    title = "{LongBench v2: Towards Deeper Understanding and Reasoning on Realistic Long-context Multitasks}",
    author = "Bai, Yushi  and
      Tu, Shangqing  and
      Zhang, Jiajie  and
      Peng, Hao  and
      Wang, Xiaozhi  and
      Lv, Xin  and
      Cao, Shulin  and
      Xu, Jiazheng  and
      Hou, Lei  and
      Dong, Yuxiao  and
      Tang, Jie  and
      Li, Juanzi",
    editor = "Che, Wanxiang  and
      Nabende, Joyce  and
      Shutova, Ekaterina  and
      Pilehvar, Mohammad Taher",
    booktitle = "Proceedings of the 63rd Annual Meeting of the Association for Computational Linguistics (Volume 1: Long Papers)",
    month = jul,
    year = "2025",
    address = "Vienna, Austria",
    publisher = "Association for Computational Linguistics",
    url = "https://aclanthology.org/2025.acl-long.183/",
    doi = "10.18653/v1/2025.acl-long.183",
    pages = "3639--3664",
    ISBN = "979-8-89176-251-0"
}

@inproceedings{ainslie-etal-2023-gqa,
    title = "{GQA: Training Generalized Multi-Query Transformer Models from Multi-Head Checkpoints}",
    author = "Ainslie, Joshua  and
      Lee-Thorp, James  and
      de Jong, Michiel  and
      Zemlyanskiy, Yury  and
      Lebron, Federico  and
      Sanghai, Sumit",
    editor = "Bouamor, Houda  and
      Pino, Juan  and
      Bali, Kalika",
    booktitle = "Proceedings of the 2023 Conference on Empirical Methods in Natural Language Processing",
    month = dec,
    year = "2023",
    address = "Singapore",
    publisher = "Association for Computational Linguistics",
    url = "https://aclanthology.org/2023.emnlp-main.298/",
    doi = "10.18653/v1/2023.emnlp-main.298",
    pages = "4895--4901"
}

@inproceedings{bai-etal-2024-longbench,
    title = "{LongBench: A Bilingual, Multitask Benchmark for Long Context Understanding}",
    author = "Bai, Yushi  and
      Lv, Xin  and
      Zhang, Jiajie  and
      Lyu, Hongchang  and
      Tang, Jiankai  and
      Huang, Zhidian  and
      Du, Zhengxiao  and
      Liu, Xiao  and
      Zeng, Aohan  and
      Hou, Lei  and
      Dong, Yuxiao  and
      Tang, Jie  and
      Li, Juanzi",
    editor = "Ku, Lun-Wei  and
      Martins, Andre  and
      Srikumar, Vivek",
    booktitle = "Proceedings of the 62nd Annual Meeting of the Association for Computational Linguistics (Volume 1: Long Papers)",
    month = aug,
    year = "2024",
    address = "Bangkok, Thailand",
    publisher = "Association for Computational Linguistics",
    url = "https://aclanthology.org/2024.acl-long.172/",
    doi = "10.18653/v1/2024.acl-long.172",
    pages = "3119--3137"
}

@inproceedings{cskv2024,
  title={{CSKV}: Training-Efficient Channel Shrinking for {KV} Cache in Long-Context Scenarios},
  author={Wang, Luning and Li, Shiyao and Ning, Xuefei and Yuan, Zhihang and Yan, Shengen and Dai, Guohao and Wang, Yu},
  booktitle={Proceedings of The 4th NeurIPS Efficient Natural Language and Speech Processing Workshop},
  pages={468--484},
  year={2024},
  volume={262},
  series={Proceedings of Machine Learning Research},
  publisher={PMLR},
  url={https://proceedings.mlr.press/v262/wang24a.html}
}

@inproceedings{kqsvd2026,
  title = 	 {{KQ-SVD: Compressing the KV Cache with Provable Guarantees on Attention Fidelity}},
  author =       {Lesens, Damien and Rakhshan, Beheshteh T. and Rabusseau, Guillaume},
  booktitle = 	 {Proceedings of The 29th International Conference on Artificial Intelligence and Statistics},
  pages = 	 {3556--3564},
  year = 	 {2026},
  editor = 	 {Khan, Emtiyaz and Li, Yingzhen and Solin, Arno and Ramdas, Aaditya},
  volume = 	 {300},
  series = 	 {Proceedings of Machine Learning Research},
  month = 	 {02--05 May},
  publisher =    {PMLR},
  url = 	 {https://proceedings.mlr.press/v300/lesens26a.html},
  address = {Tangier, Morocco}
}

@inproceedings{starkv2026,
title={{STAR-KV: Low-Rank KV Cache Compression via Soft Thresholding for Adaptive Rank Control}},
author={Priyansh Bhatnagar and Ashkan Moradifirouzabadi and Se-Hyun Yang and SeungJae Lee and Jungwook Choi and Mingu Kang},
booktitle={Forty-third International Conference on Machine Learning},
year={2026},
url={https://openreview.net/forum?id=lJjH1q6RwY},
address = {Seoul, South Korea}
}

@inproceedings{rocketkv2025,
  title = 	 {{RocketKV: Accelerating Long-Context LLM Inference via Two-Stage KV Cache Compression}},
  author =       {Behnam, Payman and Fu, Yaosheng and Zhao, Ritchie and Tsai, Po-An and Yu, Zhiding and Tumanov, Alexey},
  booktitle = 	 {Proceedings of the 42nd International Conference on Machine Learning},
  pages = 	 {3358--3392},
  year = 	 {2025},
  editor = 	 {Singh, Aarti and Fazel, Maryam and Hsu, Daniel and Lacoste-Julien, Simon and Berkenkamp, Felix and Maharaj, Tegan and Wagstaff, Kiri and Zhu, Jerry},
  volume = 	 {267},
  series = 	 {Proceedings of Machine Learning Research},
  month = 	 {13--19 Jul},
  publisher =    {PMLR},
  url = 	 {https://proceedings.mlr.press/v267/behnam25a.html},
  address = {Vancouver, Canada}
}

@inproceedings{minicache2024,
 author = {Liu, Akide and Liu, Jing and Pan, Zizheng and He, Yefei and Haffari, Gholamreza and Zhuang, Bohan},
 booktitle = {Advances in Neural Information Processing Systems},
 doi = {10.52202/079017-4443},
 editor = {A. Globerson and L. Mackey and D. Belgrave and A. Fan and U. Paquet and J. Tomczak and C. Zhang},
 pages = {139997--140031},
 publisher = {Curran Associates, Inc.},
 title = {{MiniCache: KV Cache Compression in Depth Dimension for Large Language Models}},
 url = {https://proceedings.neurips.cc/paper_files/paper/2024/file/fd0705710bf01b88a60a3d479ea341d9-Paper-Conference.pdf},
 volume = {37},
 year = {2024},
 address = {Vancouver, Canada},
}

@inproceedings{zhang-etal-2025-leank,
    title = "{LeanK: Learnable K Cache Channel Pruning for Efficient Decoding}",
    author = "Zhang, Yike  and
      He, Zhiyuan  and
      Jiang, Huiqiang  and
      Zhang, Chengruidong  and
      Yang, Yuqing  and
      Wang, Jianyong  and
      Qiu, Lili",
    editor = "Christodoulopoulos, Christos  and
      Chakraborty, Tanmoy  and
      Rose, Carolyn  and
      Peng, Violet",
    booktitle = "Proceedings of the 2025 Conference on Empirical Methods in Natural Language Processing",
    month = nov,
    year = "2025",
    address = "Suzhou, China",
    publisher = "Association for Computational Linguistics",
    url = "https://aclanthology.org/2025.emnlp-main.1584/",
    doi = "10.18653/v1/2025.emnlp-main.1584",
    pages = "31122--31137",
    ISBN = "979-8-89176-332-6"
}

\appendix

\section{Preliminaries}
\label{sec:appendix_preliminaries}

\subsection{KV Cache Computation and Update}
The inference process of typical LLMs consists of two main stages: \textit{prefilling} and \textit{decoding}.
During the \textit{prefilling} stage, the input prompt is passed through all layers of the model to generate the KV cache for each transformer layer. This cache stores the key and value tensors, which are reused in subsequent \textit{decoding} steps, significantly reducing redundant computation.
Additionally, during the \textit{decoding} stage, each newly generated token is also projected into key and value tensors, which are appended to the existing KV cache. In this way, the KV cache is progressively updated as the sequence grows, facilitating efficient autoregressive generation.

\paragraph{Prefilling} Let $\mathbf{X}\in\mathbb{R}^{\mathcal{B}\times \mathcal{T}\times \mathcal{D}}$ denote the input tensor, where $\mathcal{B}$ is the batch size, $\mathcal{T}$ is the prompt length, and $\mathcal{D}$ is the hidden dimension. The corresponding key and value tensors, $\mathbf{S}^{K}$ and $\mathbf{S}^{V}$, are computed as follows:
\begin{equation}
  \mathbf{S}^{K/V} = \mathbf{X} \mathbf{W}^{K/V},
\end{equation}
where $\mathbf{W}^{K/V} \in \mathbb{R}^{\mathcal{D} \times \mathcal{D}'}$ are the key and value projection matrices, respectively. Here, $\mathcal{D}' = h_{kv} \cdot d_h$, with $h_{kv}$ representing the number of KV heads and $d_h$ the head dimension. The computed $\mathbf{S}^{K/V}\in\mathbb{R}^{\mathcal{B}\times\mathcal{T}\times\mathcal{D}'}$ are stored as the initial KV cache to enable efficient retrieval during \textit{decoding}.

Similarly, the query tensor is obtained as
\begin{equation}
  \mathbf{Q} = \mathbf{X}\mathbf{W}^{Q},
\end{equation}
where $\mathbf{W}^{Q}\in\mathbb{R}^{\mathcal{D}\times \mathcal{G}\mathcal{D}'}$. The parameter $\mathcal{G}$ denotes the query-to-KV head ratio. When $\mathcal{G}>1$, the model corresponds to grouped-query attention (GQA); when $\mathcal{G}=1$, it reduces to multi-head attention (MHA).

\paragraph{Decoding} At each \textit{decoding} timestep $t$, the newly generated input token  $\mathbf{x}_t \in \mathbb{R}^{\mathcal{B} \times 1 \times \mathcal{D}}$ is projected to obtain the corresponding key and value vectors $\mathbf{s}^{K/V}_t$:
\begin{equation}
  \mathbf{s}^{K/V}_t = \mathbf{x}_t \mathbf{W}^{K/V}.
\end{equation}
The KV cache is then updated by appending the new entries:
\begin{equation}
  \mathbf{S}^{K/V}_{t} = \mathbf{S}^{K/V}_{t-1} \oplus \mathbf{s}^{K/V}_t,
  \label{eq:decoding_update}
\end{equation}
where $\oplus$ denotes concatenation along $\mathcal{T}$.

\subsection{Singular Value Decomposition}

SVD is a fundamental tool in matrix analysis that enables us to decompose any matrix into a set of orthogonal basis vectors and singular values, facilitating an understanding of its structure and enabling efficient computations such as dimensionality reduction. Given a matrix $\mathbf{A} \in \mathbb{R}^{m \times n}$, the SVD factorizes it as follows:
\begin{equation}
  \text{SVD}(\mathbf{A}) = \mathbf{U} \mathbf{\Sigma} \mathbf{Z}^\top,
\end{equation}
where $\mathbf{U} \in \mathbb{R}^{m \times m}$ and $\mathbf{Z} \in \mathbb{R}^{n \times n}$ are orthogonal matrices whose columns are, respectively, the left and right singular vectors of $\mathbf{A}$. The matrix $\mathbf{\Sigma} \in \mathbb{R}^{m \times n}$ is diagonal (typically rectangular), with non-negative singular values arranged in descending order.

For practical purposes such as dimensionality reduction and compression, one often uses the truncated rank-$r$ SVD, which retains only the $r$ largest singular values and their corresponding singular vectors:
\begin{equation}
  \text{SVD}_r(\mathbf{A}) = \mathbf{U}_r \mathbf{\Sigma}_r \mathbf{Z}_r^\top,
\end{equation}
where $\mathbf{U}_r \in \mathbb{R}^{m \times r}$ and $\mathbf{Z}_r \in \mathbb{R}^{n \times r}$ contain the first $r$ left and right singular vectors, respectively, and $\mathbf{\Sigma}_r \in \mathbb{R}^{r \times r}$ is a diagonal matrix of the top $r$ singular values.

\section{Decoding-Time Complexity Comparison}
\label{sec:appendix_decoding_complexity}

This section provides a detailed decoding-time complexity comparison between S$^4$R and a full-reconstruction low-rank baseline.
We focus on the per-layer cost at a single decoding step and omit the batch dimension for clarity.
Let $\mathcal{L}_t$ denote the current non-sink cache length, $s$ the number of sink tokens, $\mathcal{D}'$ the KV hidden dimension, $r$ the low-rank dimension, $\mathcal{G}$ the query-to-KV head ratio, $\mathcal{W}$ the query-window size, and $\rho$ the sparse reconstruction ratio.
The number of non-sink tokens reconstructed by S$^4$R is
\begin{equation}
  m_t = |\widehat{\mathcal{I}}|
  = \max(\mathcal{W}, \lfloor \rho \mathcal{L}_t \rfloor).
\end{equation}
For long contexts where $\rho\mathcal{L}_t \geq \mathcal{W}$, we have $m_t \approx \rho\mathcal{L}_t$.

\subsection{Full-Reconstruction Low-Rank Decoding}

A natural alternative to sparse reconstruction is to store the non-sink KV cache in coefficient form but reconstruct the entire cache at every decoding step before applying attention.
In this full-reconstruction baseline, all cached non-sink keys and values are recovered as
\begin{equation}
  \mathbf{K}_t = \mathbf{C}_t^K(\mathbf{B}^K)^\top,
  \qquad
  \mathbf{V}_t = \mathbf{C}_t^V(\mathbf{B}^V)^\top.
\end{equation}
The reconstruction cost for keys and values is therefore
\begin{equation}
  O(2\mathcal{L}_t r\mathcal{D}'),
\end{equation}
which we write as $O(\mathcal{L}_t r\mathcal{D}')$ after dropping constant factors.
The final attention is then computed over the full non-sink cache, together with the sink tokens.
Ignoring the small sink-token term for long contexts, the attention score computation and value aggregation have cost
\begin{equation}
  O(\mathcal{G}\mathcal{L}_t\mathcal{D}').
\end{equation}
Thus, the dominant per-step decoding complexity of full reconstruction is
\begin{equation}
  \begin{aligned}
    C_{\mathrm{full\text{-}rec}}
    &= O\left(\mathcal{L}_t r\mathcal{D}'
    + \mathcal{G}\mathcal{L}_t\mathcal{D}'\right)\\
    &= O\left(\mathcal{L}_t\mathcal{D}'(r+\mathcal{G})\right).
  \end{aligned}
  \label{eq:full_reconstruction_complexity}
\end{equation}
This baseline reduces persistent KV memory because it stores low-rank coefficients, but its decoding computation still scales with the full cache length in both reconstruction and attention.

\subsection{S$^4$R Sparse-Reconstruction Decoding}

S$^4$R avoids reconstructing the full cache.
At each decoding step, the new query, key, and value states are first projected into the learned low-rank subspaces.
The cost of this cache update is
\begin{equation}
  O((\mathcal{G}+2)\mathcal{D}'r),
\end{equation}
where the $\mathcal{G}$ term comes from grouped query projection and the remaining two terms come from key and value projection.
This term does not scale with $\mathcal{L}_t$ and is therefore lower order for long contexts.

S$^4$R then estimates token relevance in the latent space by multiplying the query-coefficient window with the cached key coefficients:
\begin{equation}
  \mathbf{Z}=\frac{\mathbf{C}^Q(\mathbf{C}_t^K)^\top}{\sqrt r},
  \qquad
  \mathbf{Z}\in\mathbb{R}^{\mathcal{G}\times\mathcal{W}\times\mathcal{L}_t}.
\end{equation}
This latent scoring step costs
\begin{equation}
  O(\mathcal{G}\mathcal{W}\mathcal{L}_t r).
\end{equation}
The subsequent softmax, pooling, group aggregation, and TopK selection add lower-order terms such as $O(\mathcal{G}\mathcal{W}\mathcal{L}_t)$ and $O(\mathcal{L}_t\log k)$, or approximately $O(\mathcal{L}_t)$ with efficient selection.

After token selection, S$^4$R reconstructs only $m_t$ selected non-sink tokens:
\begin{equation}
  \widehat{\mathbf{K}}
  = \mathbf{C}_t^K{}_{[\widehat{\mathcal{I}}]}(\mathbf{B}^K)^\top,
  \qquad
  \widehat{\mathbf{V}}
  = \mathbf{C}_t^V{}_{[\widehat{\mathcal{I}}]}(\mathbf{B}^V)^\top.
\end{equation}
The sparse reconstruction cost is
\begin{equation}
  O(2m_t r\mathcal{D}'),
\end{equation}
which is $O(m_t r\mathcal{D}')$ after omitting constant factors.
The final exact attention is performed only over the reconstructed non-sink tokens and the full-precision sink tokens, with cost
\begin{equation}
  O(\mathcal{G}(s+m_t)\mathcal{D}').
\end{equation}
Ignoring lower-order terms and the small sink-token contribution, the dominant S$^4$R decoding complexity is therefore
\begin{equation}
  C_{\mathrm{S^4R}}
  = O\left(
  \mathcal{G}\mathcal{W}\mathcal{L}_t r
  + m_t r\mathcal{D}'
  + \mathcal{G}m_t\mathcal{D}'
  \right).
  \label{eq:s4r_decoding_complexity_general}
\end{equation}
When $m_t\approx \rho\mathcal{L}_t$, this becomes
\begin{equation}
  C_{\mathrm{S^4R}}
  = O\left(
  \mathcal{G}\mathcal{W}\mathcal{L}_t r
  + \rho\mathcal{L}_t r\mathcal{D}'
  + \rho\mathcal{G}\mathcal{L}_t\mathcal{D}'
  \right).
  \label{eq:s4r_decoding_complexity_rho}
\end{equation}
Equivalently,
\begin{equation}
  C_{\mathrm{S^4R}}
  = O\left(
  \mathcal{L}_t\left[
  \mathcal{G}\mathcal{W}r
  + \rho\mathcal{D}'(r+\mathcal{G})
  \right]
  \right).
  \label{eq:s4r_decoding_complexity_compact}
\end{equation}

\subsection{Asymptotic Comparison}

Equations~\eqref{eq:full_reconstruction_complexity} and~\eqref{eq:s4r_decoding_complexity_compact} show the main difference between the two approaches.
Full reconstruction pays the reconstruction and attention cost over all $\mathcal{L}_t$ cached non-sink tokens;
S$^4$R reduces both reconstruction and exact attention from $\mathcal{L}_t$ tokens to approximately $\rho\mathcal{L}_t$ tokens, but introduces latent-space relevance scoring.
Therefore, S$^4$R is computationally favorable over full reconstruction when
\begin{equation}
  \mathcal{G}\mathcal{W}r
  + \rho\mathcal{D}'(r+\mathcal{G})
  <
  \mathcal{D}'(r+\mathcal{G}).
\end{equation}
Equivalently,
\begin{equation}
  \mathcal{G}\mathcal{W}r
  <
  (1-\rho)\mathcal{D}'(r+\mathcal{G}).
  \label{eq:s4r_beneficial_condition}
\end{equation}
This condition is usually satisfied when the query window $\mathcal{W}$ and rank $r$ are much smaller than the original KV dimension $\mathcal{D}'$, and when the reconstruction ratio $\rho$ is substantially below one.

\subsection{Concrete Example}

We instantiate the above comparison using the latency-oriented efficiency setting in Table~\ref{tab:efficiency}, where $\mathcal{W}=4$.
Consider $r=204$, $\rho=0.2$, and $\mathcal{D}'=1024$.
After removing the common factor $\mathcal{L}_t$, the full-reconstruction cost coefficient is
\begin{equation}
  \begin{aligned}
    r\mathcal{D}' + \mathcal{G}\mathcal{D}'
    &= 204\times1024 + 1024\mathcal{G}\\
    &= 208{,}896 + 1024\mathcal{G}.
  \end{aligned}
\end{equation}
For S$^4$R, the latent-scoring coefficient is
\begin{equation}
  \mathcal{G}\mathcal{W}r
  = 4\times204\mathcal{G}
  = 816\mathcal{G},
\end{equation}
while the sparse reconstruction coefficient is
\begin{equation}
  \rho r\mathcal{D}'
  = 0.2\times204\times1024
  = 41{,}779.2,
\end{equation}
 and the sparse attention coefficient is
\begin{equation}
  \rho\mathcal{G}\mathcal{D}'
  = 0.2\times1024\mathcal{G}
  = 204.8\mathcal{G}.
\end{equation}
Thus,
\begin{equation}
  C_{\mathrm{S^4R}} / \mathcal{L}_t
  \approx 41{,}779 + 1021\mathcal{G},
\end{equation}
whereas
\begin{equation}
  C_{\mathrm{full\text{-}rec}} / \mathcal{L}_t
  \approx 208{,}896 + 1024\mathcal{G}.
\end{equation}
For $\mathcal{G}=1$, this gives approximately
\begin{equation}
  \frac{C_{\mathrm{S^4R}}}{C_{\mathrm{full\text{-}rec}}}
  \approx
  \frac{42{,}800}{209{,}920}
  \approx 0.204.
\end{equation}
For $\mathcal{G}=4$, the ratio is
\begin{equation}
  \frac{C_{\mathrm{S^4R}}}{C_{\mathrm{full\text{-}rec}}}
  \approx
  \frac{45{,}862}{212{,}922}
  \approx 0.215.
\end{equation}
These estimates indicate that, under this setting, S$^4$R reduces the dominant reconstruction-and-attention arithmetic by roughly $4.6$--$5\times$ relative to full reconstruction.
The latent scoring term is small compared with the saved reconstruction cost because
\begin{equation}
  \mathcal{W}r = 4\times204=816
  \ll
  \rho r\mathcal{D}' = 41{,}779.2.
\end{equation}
Consequently, the leading effect is that sparse reconstruction replaces the full reconstruction cost $O(\mathcal{L}_t r\mathcal{D}')$ with $O(\rho\mathcal{L}_t r\mathcal{D}')$, yielding an approximate $1/\rho$ reconstruction-compute reduction when the latent-scoring overhead is small.

\section{More Related Work}
\label{sec:appendix_related_work}

The main text positions S$^4$R within low-rank KV-cache compression. This appendix covers complementary compression axes and then isolates the operational differences between S$^4$R and its nearest alternatives.

\subsection{Complementary KV-Cache Compression Axes}
\label{sec:appendix_beyond_low_rank_compression}

\paragraph{Sequence reduction and query-aware retrieval.}
Eviction methods reduce persistent memory by retaining only a subset of past positions. StreamingLLM combines attention sinks with a sliding window, 
while H2O and Scissorhands retain tokens using accumulated or persistent importance signals \cite{efficient2024,h2o2023,scissorhands2023}. 
Later methods allocate the token budget more selectively: NACL combines local and global importance, 
Ada-KV distributes cache budgets across heads, 
SnapKV uses recent prompt attention, 
and PyramidInfer and PyramidKV vary retention across layers \cite{nacl2024,adakv2025,snapkv2024,yang-etal-2024-pyramidinfer,pyramidkv2025}. 
Quest and DuoAttention make access query- or head-dependent rather than applying a uniform retention policy \cite{quest2024,duoattention2025}. 
RocketKV combines coarse permanent eviction with fine-grained top-$k$ sparse attention, using both head- and sequence-dimensional approximations during selection \cite{rocketkv2025}. 
These approaches principally reduce the sequence dimension or the set of KV entries accessed by attention. 
S$^4$R instead keeps a coefficient representation for every non-sink position and makes the full-dimensional reconstruction set query-dependent, so positions omitted at one decoding step remain recoverable later.

\paragraph{Token- and layer-level merging.}
Merging methods preserve information from removed entries by combining them with retained representatives. 
CaM merges to-be-evicted states according to attention-derived importance, 
KVMerger identifies groups of similar states, 
Chelsea performs online clustering along the sequence dimension, 
and GRKV distributes information through a global regression objective \cite{cam,model2024,dynakv2025,peng2026grkv}. 
MiniCache exploits a different redundancy axis by merging KV states across adjacent middle-to-deep layers \cite{minicache2024}. 
These methods reduce the number of independently stored states across tokens or layers. 
S$^4$R does not merge token identities or layer states: 
it stores per-position coefficients in layer-specific key/value subspaces and reconstructs selected positions when needed. 
Token/layer merging and coefficient-space compression are therefore distinct and potentially composable, 
but they impose different information-loss mechanisms.

\paragraph{Quantization and channel pruning.}
Quantization reduces the bit width of cached states. 
KIVI uses asymmetric low-bit schemes for keys and values, 
KVQuant combines per-channel and non-uniform quantization with explicit outlier handling, 
and GEAR approximates quantization error with low-rank and sparse residual components \cite{kivi2023,kvquant,gear2024}. 
TaDA further adapts precision across layers with mean-centering \cite{joshi-etal-2025-tada}. 
LeanK targets the channel dimension directly by learning static sparsity for the key cache \cite{zhang-etal-2025-leank}. 
In contrast, S$^4$R represents both keys and values through prompt-derived low-rank bases and coefficient caches. Bit-width reduction or channel pruning could be applied to those coefficients, but such combinations would require evaluating whether quantization or pruning distorts latent-space relevance scores as well as reconstructed attention.

\paragraph{Memory hierarchy and cache reuse.}
System-oriented methods change where compressed information resides or how it is reused. 
KV-Compress couples block eviction with a PagedAttention-style memory manager, 
KVzip builds query-agnostic compressed caches reusable across future queries, 
and SpeCache keeps the full cache in CPU memory while using a compact GPU-side representation to guide fetching \cite{kvcompress2024,kvzip2025,specache2025}. 
These methods address allocation, reuse, and data movement, 
whereas S$^4$R changes the mathematical representation stored for each prompt. 
Applying such placement or offloading policies to a coefficient cache is a systems-level extension rather than a competing subspace-construction method.

\subsection{Fine-Grained Positioning of S$^4$R}

\paragraph{Where and when the low-rank representation is constructed.}
Palu and LoRC determine their factorized representations before serving the current prompt, 
avoiding prompt-time decomposition but not adapting the subspace to the realized context \cite{palu2025,lorc2024}. 
ShadowKV and xKV construct prompt-dependent representations and are therefore closer to S$^4$R \cite{shadowkv2025,xkv2025}. 
ShadowKV decomposes current-prompt pre-RoPE keys, 
keeps a low-rank key representation on GPU, 
and retrieves selected values from an offloaded cache; 
xKV performs prompt-time cross-layer SVD and shares aligned low-rank structure across grouped layers. 
S$^4$R instead estimates separate key and value bases from a sampled subset of non-sink prompt tokens, 
then stores coefficients for all non-sink positions. 
The intended trade-off is narrower: 
it preserves prompt dependence while reducing the amount of prompt data participating in decomposition. 
Our efficiency measurements, rather than the method definitions alone, 
establish the resulting TTFT and generation-latency differences.

\paragraph{Persistent compression versus sparse access.}
StreamingLLM and SnapKV permanently retain a selected subset of full-dimensional positions, 
while Quest and Loki primarily reduce attention work through query-dependent sparse access \cite{efficient2024,snapkv2024,quest2024,loki2024}. 
RocketKV explicitly combines permanent eviction with a second sparse-attention stage \cite{rocketkv2025}. 
S$^4$R couples these concerns differently: 
its persistent memory reduction comes from coefficient-space storage, 
whereas latent scoring determines only which positions are reconstructed for the current attention step. 
This separation lets a position remain cheaply recoverable after it is skipped, 
at the cost of storing coefficients for the full non-sink sequence and performing latent scoring during decoding.

\paragraph{Takeaway.}
Across these axes, the distinctive combination in S$^4$R is prompt-sampled construction of both key and value subspaces, persistent per-position coefficient storage, full-precision sink preservation, and query-dependent sparse reconstruction. Fixed low-rank methods remove most prompt-time adaptation; prompt-wide methods retain adaptation but pay decomposition or update costs over more prompt states; eviction and retrieval methods reduce the active sequence without necessarily compressing every retained state; and quantization, merging, and systems techniques target complementary representation or deployment dimensions.

\section{More Ablation Studies}
\label{sec:appendix_more_ablations}

\subsection{Effect of Query Window}
\label{sec:appendix_effect_query_window}

\begin{table}[tb]
  \centering
  \small
  \setlength{\tabcolsep}{4.5pt}
  \begin{tabular}{ccccccc}
    \toprule
    $\mathcal{W}$ & Fewshot & SDQA & MDQA & LCC & PRE & Avg. \\
    \midrule
    \multicolumn{7}{c}{Llama-3.2-1B}\\
    $1$ & 46.82 & 20.53 & 21.13 & 19.41 & 4.15 & 25.90 \\
    $4$ & 52.60 & 19.99 & 22.60 & 23.56 & 3.95 & 28.46 \\
    $8$ & 52.71 & 20.13 & 22.75 & 23.41 & 3.95 & 28.56 \\
    $16$ & 53.32 & 20.22 & 22.26 & 24.25 & 3.86 & 28.68 \\
    \rowcolor{Salmon!40} $32$ & 51.29 & 20.69 & 22.06 & 23.43 & 3.39 & 28.09 \\
    $64$ & 48.82 & 20.60 & 21.80 & 23.84 & 3.72 & 27.38 \\
    \midrule
    \multicolumn{7}{c}{Qwen3-4B}\\
    $1$ & 64.83 & 37.87 & 37.40 & 5.08 & 95.00 & 47.31 \\
    $4$ & 65.05 & 38.37 & 37.98 & 4.99 & 95.83 & 47.73 \\
    $8$ & 64.99 & 38.55 & 37.81 & 5.01 & 95.33 & 47.67 \\
    $16$ & 65.79 & 38.56 & 37.63 & 5.03 & 95.33 & 47.85 \\
    \rowcolor{Salmon!40} $32$ & 65.40 & 38.53 & 37.56 & 5.05 & 96.17 & 47.79 \\
    $64$ & 65.30 & 38.11 & 37.56 & 5.11 & 95.00 & 47.55 \\
    \midrule
    \multicolumn{7}{c}{Llama-3.1-8B}\\
    $1$ & 61.62 & 41.13 & 44.64 & 29.49 & 99.50 & 51.92 \\
    $4$ & 63.60 & 42.38 & 46.33 & 32.06 & 99.50 & 53.50 \\
    $8$ & 62.99 & 42.74 & 45.98 & 32.14 & 99.50 & 53.34 \\
    $16$ & 62.16 & 42.75 & 46.08 & 32.22 & 99.50 & 53.15 \\
    \rowcolor{Salmon!40} $32$ & 61.92 & 42.32 & 46.28 & 32.29 & 99.50 & 53.03 \\
    $64$ & 58.75 & 42.03 & 46.24 & 32.46 & 99.50 & 52.09 \\
    \midrule
    \multicolumn{7}{c}{Qwen2.5-14B}\\
    $1$ & 66.16 & 38.05 & 52.16 & 19.38 & 92.50 & 52.82 \\
    $4$ & 67.01 & 38.91 & 51.99 & 19.97 & 94.50 & 53.47 \\
    $8$ & 66.88 & 38.98 & 52.59 & 20.19 & 94.50 & 53.64 \\
    $16$ & 67.19 & 38.65 & 52.69 & 20.29 & 94.50 & 53.67 \\
    \rowcolor{Salmon!40} $32$ & 66.71 & 38.20 & 51.88 & 20.25 & 94.50 & 53.19 \\
    $64$ &64.08 & 38.85 & 51.51 & 20.45 & 93.50 & 52.48 \\
    \bottomrule
  \end{tabular}
  \caption{Effect of the query-window size $\mathcal{W}\in\{1,4,8,16,32,64\}$ on grouped LongBench performance. All other S$^4$R hyperparameters are fixed, and shaded rows denote the default setting used in the main experiments.
  }
  \label{tab:query_window_comparison}
\end{table}

S$^4$R uses a sliding query window of length $\mathcal{W}$ for two related purposes during decoding.
First, the projected queries in this window are used to estimate token relevance in the latent coefficient space, where attention scores are averaged across recent queries before selecting globally important positions.
Second, the same window determines the local tokens that are always reconstructed and attended to in full precision.
Thus, $\mathcal{W}$ controls how much short-range decoding context is used for stable token scoring and how much recent context is guaranteed to be preserved in the sparse reconstruction set.

This ablation varies $\mathcal{W}$ while keeping the rank, sampling strategy, reconstruction ratio, and pooling configuration fixed.
When $\mathcal{W}=1$, token selection is driven only by the current query, which provides the lowest query-coefficient storage and latent scoring cost but may produce noisier global-token estimates.
Larger windows aggregate relevance over multiple recent queries and preserve more local tokens, which can improve robustness for tasks where the answer depends on a short span of recent decoding context.
However, increasing $\mathcal{W}$ also increases the query-coefficient buffer size and the latent attention cost, and it allocates more of the reconstruction budget to local tokens, leaving fewer slots for globally selected positions when the total reconstruction ratio is fixed.

Table~\ref{tab:query_window_comparison} compares $\mathcal{W}\in\{1,4,8,16,32,64\}$ across LongBench task groups.
The results show a clear preference for moderate query windows.
Using only the current query is consistently weaker: increasing $\mathcal{W}$ from $1$ to the best-performing setting improves the average score from 25.90 to 28.68 on Llama-3.2-1B, from 47.31 to 47.85 on Qwen3-4B, from 51.92 to 53.50 on Llama-3.1-8B, and from 52.82 to 53.67 on Qwen2.5-14B.
These gains indicate that aggregating relevance over several recent queries produces more stable global-token estimates than single-query selection.

The benefit, however, saturates quickly.
The strongest averages are obtained by $\mathcal{W}=16$ for Llama-3.2-1B, Qwen3-4B, and Qwen2.5-14B, and by $\mathcal{W}=4$ for Llama-3.1-8B.
The default $\mathcal{W}=32$ remains close to the optimum, trailing the best setting by 0.59, 0.06, 0.47, and 0.48 points on the four models, respectively, while maintaining a larger guaranteed local reconstruction window.
Increasing the window further to $\mathcal{W}=64$ lowers the average score for all models compared with $\mathcal{W}=32$.
This suggests that overly large query windows dilute current-step relevance and reserve more reconstruction budget for local tokens, thereby reducing the coverage of globally important positions.
We therefore keep $\mathcal{W}=32$ as the default in the main experiments not because it is always the best setting in this ablation, but because it provides near-optimal accuracy and matches the default window size used by SnapKV.

\subsection{Effect of Different Rank Allocation}
\label{sec:appendix_different_rank_allocation}

\begin{table}[tb]
  \centering
  \small
  \setlength{\tabcolsep}{3pt}
  \begin{tabular}{ccccccc}
    \toprule
    $r^K:r^V$ & Fewshot & SDQA & MDQA & LCC & PRE & Avg. \\
    \midrule
    \multicolumn{7}{c}{Llama-3.2-1B}\\
    \rowcolor{Salmon!40} $1 : 1$ & 51.29 & 20.69 & 22.06 & 23.43 & 3.39 & 28.09 \\
    $1 : 1.5$ & 51.24 & 20.46 & 22.15 & 23.44 & 4.88 & 28.17 \\
    $1 : 2$ & 50.17 & 20.33 & 22.87 & 24.33 & 5.06 & 28.14  \\
    \midrule
    \multicolumn{7}{c}{Qwen3-4B}\\
    \rowcolor{Salmon!40} $1 : 1$ & 65.40 & 38.53 & 37.56 & 5.05 & 96.17 & 47.79 \\
    $1 : 1.5$ & 64.62 & 39.00 & 38.81 & 5.10 & 96.50 & 48.08 \\
    $1 : 2$ & 64.74 & 38.70 & 38.57 & 5.10 & 96.83 & 48.00 \\
    \midrule
    \multicolumn{7}{c}{Llama-3.1-8B}\\
    \rowcolor{Salmon!40} $1 : 1$ & 61.92 & 42.32 & 46.28 & 32.29 & 99.50 & 53.03 \\
    $1 : 1.5$ & 60.32 & 43.42 & 46.34 & 32.07 & 99.50 & 52.89 \\
    $1 : 2$ & 59.96 & 43.13 & 45.98 & 32.39 & 99.50 & 52.65 \\
    \midrule
    \multicolumn{7}{c}{Qwen2.5-14B}\\
    \rowcolor{Salmon!40} $1 : 1$ & 66.71 & 38.20 & 51.88 & 20.25 & 94.50 & 53.19 \\
    $1 : 1.5$ & 67.03 & 39.94 & 52.29 & 20.67 & 96.50 & 54.09 \\
    $1 : 2$ & 67.13 & 39.34 & 52.02 & 20.98 & 95.50 & 53.81\\
    \bottomrule
  \end{tabular}
  \caption{Effect of key--value rank allocation under a fixed total coefficient budget. The ratio $r^K:r^V$ specifies the key and value ranks, and shaded rows denote the balanced default allocation.
  }
  \label{tab:rank_allocation_comparison}
\end{table}

Prior work observes that pre-RoPE keys exhibit stronger low-rank characteristics than values \cite{lee-etal-2025-tale,shadowkv2025}.
This asymmetry suggests that keys may tolerate a smaller latent rank, while values may require a larger rank to preserve the information aggregated into the final attention output.
Motivated by this observation, we restrict this ablation to settings where the key rank is no larger than the value rank, i.e., $r^K \leq r^V$.
The goal is to test whether shifting part of the coefficient budget from keys to values improves reconstruction quality without substantially degrading latent-space token scoring.

To study this trade-off, we compare several key-to-value rank ratios while keeping the total coefficient budget fixed.
Concretely, for a target budget corresponding to the default rank $r$, we choose $(r^K,r^V)$ such that $r^K+r^V=2r$ and vary the allocation ratio $r^K:r^V$.

Table~\ref{tab:rank_allocation_comparison} reports the results across LongBench task groups.
The results show that asymmetric rank allocation can be beneficial, but the effect depends on model scale.
On Llama-3.2-1B, the average scores are nearly tied, with $1{:}1.5$ slightly outperforming $1{:}1$ by 0.08 points and $1{:}2$ by 0.03 points.
On Qwen3-4B, allocating more rank to values improves the average score from 47.79 to 48.08 for $1{:}1.5$ and 48.00 for $1{:}2$.
The same trend is clearer on Qwen2.5-14B, where $1{:}1.5$ improves the average score from 53.19 to 54.09 and $1{:}2$ reaches 53.81.
In contrast, Llama-3.1-8B favors the balanced allocation, with $1{:}1$ achieving 53.03 compared with 52.89 and 52.65 for $1{:}1.5$ and $1{:}2$.
Overall, these results support the motivation that values can require more rank than pre-RoPE keys, but an overly value-heavy allocation is not uniformly optimal.
We therefore keep the balanced allocation as the default in the main experiments for simplicity and robustness, while noting that a mildly value-biased allocation such as $1{:}1.5$ can further improve some model families.

\subsection{Effect of Different Recent-to-Uniform Sampling Ratios}
\label{sec:appendix_different_recent_to_uniform_sampling_ratios}

\begin{table}[tb]
  \centering
  \small
  \setlength{\tabcolsep}{3pt}
  \begin{tabular}{ccccccc}
    \toprule
    Ratio & Fewshot & SDQA & MDQA & LCC & PRE & Avg. \\
    \midrule
    \multicolumn{7}{c}{Llama-3.2-1B}\\
    \rowcolor{Salmon!40} $7:3$ & 51.29 & 20.69 & 22.06 & 23.43 & 3.39 & 28.09 \\
    $6:4$ & 52.09 & 20.67 & 22.59 & 22.94 & 3.85 & 28.44 \\
    $5:5$ & 50.46 & 20.17 & 21.56 & 23.24 & 4.25 & 27.64 \\
    $4:6$ & 50.52 & 20.69 & 21.76 & 24.05 & 3.94 & 27.90 \\
    $10:0$ & 50.80 & 19.74 & 22.16 & 24.09 & 3.88 & 27.82 \\
    \midrule
    \multicolumn{7}{c}{Qwen3-4B}\\
    \midrule
    \rowcolor{Salmon!40} $7:3$ & 65.40 & 38.53 & 37.56 & 5.05 & 96.17 & 47.79 \\
    $6:4$ & 64.86 & 38.41 & 37.82 & 5.11 & 90.42 & 47.16 \\
    $5:5$ & 66.15 & 38.41 & 38.01 & 5.13 & 93.00 & 47.80 \\
    $4:6$ & 65.19 & 38.04 & 37.88 & 5.04 & 93.50 & 47.44 \\
    $10:0$ & 64.95 & 36.91 & 37.32 & 5.10 & 96.33 & 47.18 \\
    \midrule
    \multicolumn{7}{c}{Llama-3.1-8B}\\
    \rowcolor{Salmon!40} $7:3$ & 61.92 & 42.32 & 46.28 & 32.29 & 99.50 & 53.03 \\
    $6:4$ & 60.96 & 42.75 & 45.81 & 31.96 & 100.00 & 52.77 \\
    $5:5$ & 60.92 & 41.98 & 45.36 & 32.48 & 100.00 & 52.48 \\
    $4:6$ & 60.09 & 43.12 & 46.19 & 32.82 & 100.00 & 52.82 \\
    $10:0$ & 61.91 & 41.24 & 45.52 & 32.44 & 100.00 & 52.59 \\
    \midrule
    \multicolumn{7}{c}{Qwen2.5-14B}\\
    \rowcolor{Salmon!40} $7:3$ & 66.71 & 38.20 & 51.88 & 20.25 & 94.50 & 53.19 \\
    $6:4$ & 66.26 & 39.36 & 52.56 & 20.25 & 94.50 & 53.57 \\
    $5:5$ & 66.53 & 39.75 & 52.44 & 20.05 & 95.00 & 53.75 \\
    $4:6$ & 66.95 & 40.17 & 52.64 & 20.52 & 91.50 & 53.75 \\
    $10:0$ & 66.89 & 38.12 & 51.08 & 20.36 & 94.25 & 52.99 \\
    \bottomrule
  \end{tabular}
  \caption{Effect of the recent-to-uniform sampling ratio used to construct prompt-aware key/value bases. The total number of sampled tokens is fixed, and shaded rows denote the default $7{:}3$ split.
  }
  \label{tab:recent_to_uniform_sampling_ratio_comparison}
\end{table}

S$^4$R constructs the prompt-aware key/value bases from a sampled subset of non-sink tokens.
The sampling policy combines a contiguous block of recent tokens with uniformly sampled earlier tokens.
These two sources are complementary: recent tokens tend to be more relevant to the immediate decoding state and better match the local distribution near generation, while uniformly sampled tokens provide broader coverage of the full prompt and reduce the risk of missing globally important context.
The default configuration uses a recent-to-uniform split of $7:3$.

Table~\ref{tab:recent_to_uniform_sampling_ratio_comparison} compares several ratios while keeping the total number of sampled tokens fixed.
The results show that S$^4$R is relatively robust to moderate changes in the split, but combining recent and uniform samples is generally preferable to using only recent tokens.
For Llama-3.2-1B, $6:4$ obtains the best average score of 28.44, while the default $7:3$ remains close at 28.09.
For Qwen3-4B, $5:5$ and $7:3$ are nearly tied, with average scores of 47.80 and 47.79, respectively; however, $7:3$ performs better on passage retrieval than $5:5$.
For Llama-3.1-8B, the default $7:3$ achieves the best average score of 53.03, outperforming the other ratios by 0.21--0.55 points.
For Qwen2.5-14B, more uniform coverage is beneficial: $5:5$ and $4:6$ both reach 53.75, compared with 53.19 for $7:3$.
In contrast, the recent-only $10:0$ setting is not consistently competitive, suggesting that relying only on the most recent sampled tokens can miss useful global prompt structure.
Overall, the results indicate that no single split is universally optimal across all models, but hybrid recent-plus-uniform sampling is important.
We keep $7:3$ as the default because it is consistently strong, especially on Llama models and passage retrieval, while maintaining a simple fixed sampling rule.

\subsection{Effect of Reconstruction Ratio}
\label{sec:appendix_effect_reconstruction_ratio}

\begin{table}[tb]
  \centering
  \small
  \setlength{\tabcolsep}{3pt}
  \begin{tabular}{ccccccc}
    \toprule
    $\rho$ & Fewshot & SDQA & MDQA & LCC & PRE & Avg. \\
    \midrule
    \multicolumn{7}{c}{Llama-3.2-1B}\\
    \rowcolor{Salmon!40} $0.2$ & 51.29 & 20.69 & 22.06 & 23.43 & 3.39 & 28.09 \\
    $0.15$ & 50.50 & 19.83 & 22.75 & 24.00 & 3.63 & 27.90 \\
    \midrule
    \multicolumn{7}{c}{Qwen3-4B}\\
    \rowcolor{Salmon!40} $0.2$ & 65.40 & 38.53 & 37.56 & 5.05 & 96.17 & 47.79 \\
    $0.15$ & 64.69 & 38.38 & 37.83 & 5.03 & 94.33 & 47.46 \\
    \midrule
    \multicolumn{7}{c}{Llama-3.1-8B}\\
    \rowcolor{Salmon!40} $0.2$ & 61.92 & 42.32 & 46.28 & 32.29 & 99.50 & 53.03 \\
    $0.15$ & 59.40 & 42.08 & 45.78 & 32.06 & 99.00 & 52.08 \\
    \midrule
    \multicolumn{7}{c}{Qwen2.5-14B}\\
    \rowcolor{Salmon!40} $0.3$ & 66.71 & 38.20 & 51.88 & 20.25 & 94.50 & 53.19 \\
    $0.15$ & 64.10 & 38.13 & 51.48 & 18.66 & 93.50 & 52.12 \\
    \bottomrule
  \end{tabular}
  \caption{Effect of the reconstruction ratio $\rho$ during sparse decoding. The coefficient cache and other hyperparameters are unchanged, and shaded rows denote the default $\rho$ for each model setting.
  }
  \label{tab:reconstruction_ratio_comparison}
\end{table}

The reconstruction ratio $\rho$ controls how many non-sink positions are reconstructed and attended to in full precision at each decoding step.
In the main experiments, we set $\rho$ to match the target compression ratio $\eta$, so the sparse reconstruction budget scales with the memory budget used by each model setting.
This ablation tests whether S$^4$R can use a smaller reconstruction set during decoding while keeping the compressed coefficient cache and other hyperparameters unchanged.
Reducing $\rho$ directly lowers the number of reconstructed key/value vectors and the size of the exact attention set, but it may also remove globally relevant positions from the final sparse attention computation.

Table~\ref{tab:reconstruction_ratio_comparison} compares the default reconstruction ratio with a smaller setting of $\rho=0.15$.
The results show that S$^4$R is reasonably robust to reconstructing fewer tokens, but the default setting remains consistently stronger.
On Llama-3.2-1B, reducing $\rho$ from $0.2$ to $0.15$ only decreases the average score from 28.09 to 27.90, and even improves MDQA, LCC, and PRE slightly.
On Qwen3-4B, the average score drops by 0.33 points, mainly due to lower Fewshot and PRE scores.
The larger Llama-3.1-8B and Qwen2.5-14B settings are more sensitive: their average scores decrease by 0.95 and 1.07 points, respectively.
This suggests that a smaller reconstruction set can preserve most of the performance when further decoding-time savings are desired, but the default $\rho=\eta$ provides a safer accuracy--efficiency trade-off, especially for stronger models whose outputs rely more on recovering a broader set of relevant context positions.

\subsection{Effect of Layer-wise Full Reconstruction}
\label{sec:appendix_whether_need_with_full_reconstruction}

SALS applies full reconstruction to selected layers based on its layer-wise analysis, specifically the first two layers and the last layer, to reduce the approximation error in layers that are more sensitive to compression \cite{sals}.
Motivated by this design, we test whether a similar layer-wise full-reconstruction strategy is also beneficial for S$^4$R.
In this ablation, ``w/'' denotes applying full reconstruction to these selected layers, while ``w/o'' denotes the default S$^4$R configuration that uses sparse reconstruction consistently across layers.
All other hyperparameters are kept unchanged.

Table~\ref{tab:full_reconstruction_comparison} shows that layer-wise full reconstruction does not consistently improve S$^4$R.
On Llama-3.2-1B, it improves the average score from 28.09 to 28.45, mainly due to gains on Fewshot, MDQA, LCC, and PRE.
However, the same strategy reduces the average score on the other three models: from 47.79 to 47.45 on Qwen3-4B, from 53.03 to 52.33 on Llama-3.1-8B, and from 53.19 to 52.94 on Qwen2.5-14B.
The drops are especially visible in MDQA and PRE for Qwen3-4B and in Fewshot and MDQA for Llama-3.1-8B.

These results suggest that the layer-wise full-reconstruction heuristic used by SALS is not directly transferable to S$^4$R.
S$^4$R already preserves sink tokens and reconstructs a query-dependent sparse set at every layer; forcing full reconstruction on fixed layers can change the balance between exact local/global token recovery and the prompt-adaptive sparse reconstruction mechanism.
It also introduces additional decoding-time reconstruction cost.
Since the accuracy gains are limited to the smallest model and are not robust across model families, we do not use layer-wise full reconstruction as the default.

\begin{table}[tb]
  \centering
  \small
  \setlength{\tabcolsep}{3pt}
  \begin{tabular}{ccccccc}
    \toprule
    FR & Fewshot & SDQA & MDQA & LCC & PRE & Avg. \\
    \midrule
    \multicolumn{7}{c}{Llama-3.2-1B}\\
    \rowcolor{Salmon!40} w/o & 51.29 & 20.69 & 22.06 & 23.43 & 3.39 & 28.09 \\
    w/ & 52.29 & 20.04 & 22.74 & 23.99 & 3.77 & 28.45 \\
    \midrule
    \multicolumn{7}{c}{Qwen3-4B}\\
    \rowcolor{Salmon!40} w/o & 65.40 & 38.53 & 37.56 & 5.05 & 96.17 & 47.79 \\
    w/ & 65.62 & 38.48 & 36.52 & 5.03 & 95.08 & 47.45 \\
    \midrule
    \multicolumn{7}{c}{Llama-3.1-8B}\\
    \rowcolor{Salmon!40} w/o & 61.92 & 42.32 & 46.28 & 32.29 & 99.50 & 53.03 \\
    w/ & 60.27 & 42.22 & 45.64 & 32.20 & 99.00 & 52.33 \\
    \midrule
    \multicolumn{7}{c}{Qwen2.5-14B}\\
    \rowcolor{Salmon!40} w/o & 66.71 & 38.20 & 51.88 & 20.25 & 94.50 & 53.19 \\
    w/ & 66.26 & 37.82 & 52.03 & 19.97 & 94.00 & 52.94 \\
    \bottomrule
  \end{tabular}
  \caption{Effect of applying layer-wise full reconstruction to selected layers. ``w/'' applies full reconstruction to the first two layers and the last layer, whereas ``w/o'' is the default sparse-reconstruction setting; shaded rows denote the default.
  }
  \label{tab:full_reconstruction_comparison}
\end{table}

\begin{table*}[tb]
  \centering
  \small
  \begin{tabular}{lclcccc}
    \toprule
    Dataset & Source & Avg len & Decoding len &  Metric & Language & \#data \\ \midrule
    \textit{Single-Document QA} &  \\
    NarrativeQA & Literature, Film & 18,409 & 128 & F1 & English & 200 \\
    Qasper & Science & 3,619 & 128 & F1 & English & 200 \\
    MultiFieldQA-en & Multi-field & 4,559 & 64 &F1 & English & 150 \\
    \midrule
    \textit{Multi-Document QA} &  \\
    HotpotQA & Wikipedia & 9,151 & 32 & F1 & English & 200 \\
    2WikiMultihopQA & Wikipedia & 4,887 & 32 & F1 & English & 200 \\
    MuSiQue & Wikipedia & 11,214 & 32 & F1 & English & 200 \\
    \midrule
    \textit{Few-shot Learning} &  \\
    TREC & Web question & 5,177 & 64 &Accuracy (CLS) & English & 200 \\
    TriviaQA & Wikipedia, Web & 8,209 & 32 & F1 & English & 200 \\
    SAMSum & Dialogue & 6,258 & 128 &Rouge-L & English & 200 \\
    \midrule
    \textit{Synthetic Task} &  \\
    PassageRetrieval-en & Wikipedia & 9,289 & 32 & Accuracy (EM) & English & 200 \\ 
    \midrule
    \textit{Code Completion} &  \\
    LCC & Github & 1,235 & 64 & Edit Sim & Python/C\#/Java & 500 \\
    \bottomrule
  \end{tabular}
  \caption{
  Summary of LongBench dataset statistics. 
  `Avg len' denotes average input length (number of words for English/code datasets).
  `Decoding len' is the number of tokens in the decoding sequence. 
  `Accuracy (CLS)' indicates classification accuracy, and `Accuracy (EM)' stands for exact match accuracy.}
  \label{tab:longbench_stats}
\end{table*}

\begin{table*}[tb]
  \centering
  \small
  \begin{tabular}{cccc}
    \toprule
    \multirow{2}{*}{Task} & \multicolumn{3}{c}{Configurations} \\ 
    \cline{2-4} 
    & Subtask-1 & Subtask-2 & Subtask-3 \\ 
    \midrule
    \multirow{4}{*}{MK-NIAH} & num\_keys = 2 & type\_key = word & type\_key = word \\
    & type\_key = word & type\_value = number & type\_value = uuid \\
    & type\_value = number & type\_haystack = essay & type\_haystack = essay \\
    & type\_haystack = essay & \textbf{line retrieval} & \textbf{KV retrieval} \\
    \midrule
    MV-NIAH & \multicolumn{3}{l}{num\_values = 2, type\_key = word, type\_value = number, type\_haystack = essay} \\ 
    \midrule
    MQ-NIAH & \multicolumn{3}{l}{num\_queries = 2, type\_key = word, type\_value = number, type\_haystack = essay} \\ 
    \midrule
    VT & \multicolumn{3}{l}{num\_chains = 1, num\_hops = 4} \\ 
    \midrule
    FWE & \multicolumn{3}{l}{$\alpha = 2.0$} \\
    \bottomrule
  \end{tabular}
  \caption{Seven task configurations in RULER.}
  \label{tab:ruler_configs}
\end{table*}

\section{Implementation Details}
\label{sec:implementation_details}

\subsection{LongBench}
%\subsubsection{Dataset Statistics}
In this section, we provide the detailed statistics of the LongBench benchmark \cite{bai-etal-2024-longbench} used in our evaluation. As shown in Table~\ref{tab:longbench_stats}, the selected datasets cover five major categories: single-document QA, multi-document QA, few-shot learning, synthetic tasks, and code completion. These tasks represent typical long-context scenarios with average input lengths ranging from approximately 1.2k to over 18k words. We also specify the decoding lengths and the primary evaluation metrics (e.g., F1 score, Rouge-L, and Exact Match) for each task to provide a comprehensive reference for our experimental setup in Section~\ref{sec:accuracy_evaluation}.

%\subsubsection{Model Details}
We evaluate our method on the LongBench benchmark using Llama-3.2-1B-Instruct, 
Llama-3.1-8B-Instruct~\cite{llama2024}, 
Qwen3-4B-Instruct-2507~\cite{yang2025qwen3technicalreport} and Qwen2.5-14B-Instruct~\cite{qwen252025}. 
Llama-3.2-1B-Instruct is a 1-billion-parameter model with a 128k-token context window, while Llama-3.1-8B-Instruct is an 8-billion-parameter model with a 128k-token context window. Qwen3-4B-Instruct-2507 is a 4-billion-parameter model with a 32k-token context window natively and 128k tokens with YaRN~\cite{peng2026yarnefficientcontextwindow}, while Qwen2.5-14B-Instruct is a 14-billion-parameter model with a 32k-token context window.
For LongBench samples exceeding 32k tokens (e.g., in NarrativeQA), we truncate the input to the rightmost 32k tokens to fit within the Qwen2.5-14B-Instruct's constraints.

\subsection{RULER}
RULER~\cite{ruler2024} is built for maximal flexibility, supporting a vast range of sequence lengths and task complexities. 
Each task allows for a combinatorial number of configuration options, as summarized in Table~\ref{tab:ruler_configs}.
In our primary experiments, we assess Qwen2.5-7B-Instruct-1M~\cite{qwen251m2025} on seven representative tasks spanning RULER's two primary categories: retrieval and multi-hop tracing. 
For each task, we prepare samples of 128k tokens and generate 20 samples to assess performance. 
Unless otherwise specified, the decoding length is set to 32 tokens.

Both LongBench and RULER are evaluated using LM-Evaluation-Harness~\cite{lm-eval-harness2024}.

\end{document}